%% file: main.tex
\documentclass[sn-basic,NameDate,pdflatex]{sn-jnl}

\makeatletter
\renewcommand{\listfont}{\@currsize} 
\makeatother

\usepackage[dvipsnames]{xcolor}
\usepackage{graphicx}
\usepackage{multirow}
\usepackage{amsmath,amssymb,amsfonts}
\usepackage{amsthm}
\usepackage{mathrsfs}
\usepackage[title]{appendix}
\usepackage{xcolor}
\usepackage{textcomp}
\usepackage{manyfoot}
\usepackage{booktabs}
\usepackage{algorithm}
\usepackage{algorithmicx}
\usepackage{algpseudocode}
\usepackage{listings}
\usepackage{todonotes}
\usepackage{bbm}
\usepackage{subcaption}
\usepackage{tikz}
\usepackage{footnote}
\usepackage{xfrac}
\usepackage{nicefrac}
\usetikzlibrary{decorations.pathreplacing}

\makeatletter
\renewcommand{\figurecaptionfont}{\small}
\renewcommand{\tablecaptionfont}{\@currsize}
\makeatother
\DeclareCaptionFont{figurecaptionfont}{\figurecaptionfont}
\newtheorem{remark}{Remark}%
\theoremstyle{definition}
\newtheorem{definition}{Definition}
\theoremstyle{theorem}
\newtheorem{theorem}{Theorem}
\newtheorem{theo}{Theorem}[section]
\newtheorem{example}[theo]{Example}

\newcommand{\E}{\mathbb{E}}

\newcommand{\1}[1]{\mathbbm{1}_{\left\{ #1 \right\}}}

\DeclareMathOperator{\CC}{CC}

\begin{document}

\title[Article Title]{Model Monitoring: A General Framework with an Application to Non-life Insurance Pricing}

\author[1]{\fnm{Alexej} \sur{Brauer} \orcid{https://orcid.org/0009-0009-4001-016X}}\email{brauer.alexej@gmail.com}
\author[1]{\fnm{Paul} \sur{Menzel}}\email{paul.menzel@tum.de}
\author[2]{\fnm{Mario V.} \sur{W\"uthrich}\email{mario.wuethrich@math.ethz.ch} \orcid{https://orcid.org/0000-0003-4035-552X}}
\affil[1]{\orgdiv{Actuarial Department},
	\orgname{Allianz Versicherungs-AG},
	\orgaddress{\city{Munich},
		\country{Germany}}}
\affil[2]{\orgname{Department of Mathematics, ETH Zurich},
	\orgaddress{\city{Zurich},
		\country{Switzerland}}}

\abstract{Maintaining the predictive performance of pricing models is challenging
	when insurance portfolios and data-generating mechanisms evolve over time.
	Focusing on non-life insurance, we adopt the concept-drift terminology from
	machine learning and distinguish virtual drift from real concept drift in an
	actuarial setting.
	Methodologically, we (i) formalize deviance loss and Murphy's score
	decomposition to assess global and local auto-calibration; (ii) study the Gini
	score as a rank-based performance measure, derive its asymptotic distribution,
	and develop a consistent bootstrap estimator of its asymptotic variance; and
	(iii) combine these results into a statistically grounded, model-agnostic
	monitoring framework that integrates a Gini-based ranking drift test with
	global and local auto-calibration tests.
	An application to a modified motor insurance portfolio with controlled
	concept-drift scenarios illustrates how the framework guides decisions on
	refitting or recalibrating pricing models.
}

\keywords{non-life insurance, actuarial pricing, concept drift, model monitoring, Gini score, Murphy decomposition, non-stationarity}
\maketitle
\textbf{Statements and Declarations:}
\noindent
\small{\textbf{Competing Interests:}
	The authors have no conflicts of interest to declare that are relevant to the
	content of this article.
}
\newpage
\input{01_Sec1_Intro.tex}
\input{01_Sec2_TechDetail_RelWork.tex}
\input{01_Sec3_Framework_for_Model_Monitoring.tex}
\input{01_Sec4_Conclusion_and_Outlook.tex}
\bibliography{my_bibliography}
\newpage
\input{01_Sec5_Appendices.tex}
\end{document}

%% file: 01_Sec1_Intro.tex
\section{Introduction}\label{sec:introduction}
\begin{figure}[H]
	\centering
	\includegraphics[clip, trim=0.5cm 5.5cm 0.5cm 0.5cm, width=\textwidth]{
		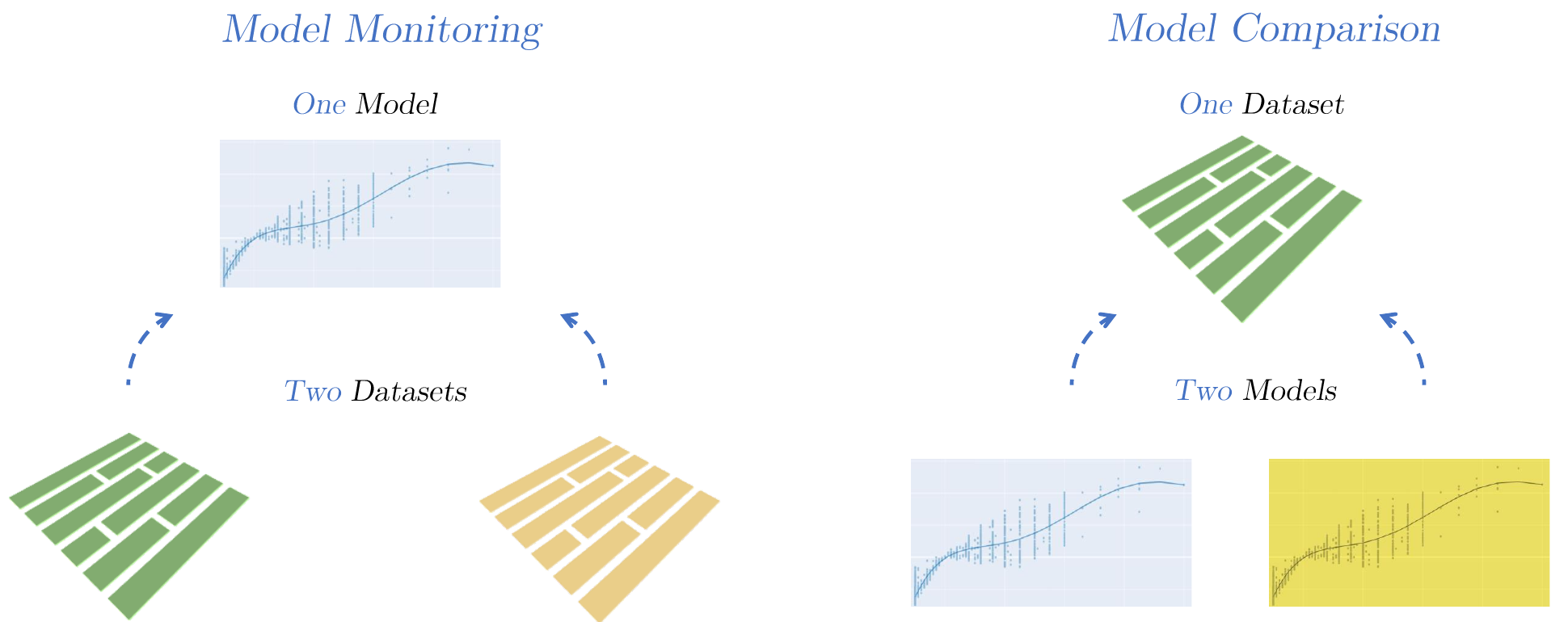}
	\caption{Schematic representation of model monitoring (left)
		and model comparison (right).\protect\footnotemark}
	\label{fig:model_monitoring_comparison}
\end{figure}

In non-life insurance pricing, common tasks include predicting claim frequency
and severity, and modeling binary demand outcomes such as conversion
prediction.
Maintaining the accuracy of those models over time is a critical challenge in a
dynamic landscape of evolving portfolios and market conditions.
In this context, two key terms are often encountered: \textit{model monitoring}
and \textit{model comparison}.
These two terms are frequently used in the actuarial context of developing
pricing models.
The term \textit{model monitoring}, also referred to as \textit{backtesting},
pertains to testing a single model on at least two different datasets.
This can occur either during the model development phase, using training and
validation/holdout data, or during the monitoring phase, using holdout data
from the model update period versus new data from the current period.
In contrast, \textit{model comparison} is the process of comparing two
different models on the same dataset to determine which one has a better
performance for a given task.
A schematic representation of both terms is given in Fig.
\ref{fig:model_monitoring_comparison}.
\footnotetext{To visually distinguish datasets, the Apache Parquet logo color has been
	modified from its original design.
	The Apache Parquet logo is a trademark of the Apache Software Foundation.
}
\noindent
In this work, we focus on \textit{model monitoring}, which is often
under-emphasized in the actuarial literature, yet it is crucial for updating and
maintaining pricing models over time.
To the best of our knowledge, this is the first work to explicitly examine data
drift in insurance pricing models; accordingly, we provide an overview of the
relevant literature and a monitoring framework tailored to the actuarial
pricing practice.
In contrast, \textit{model comparison} is a more static process, typically
performed during development to select the best model from a set of candidates
(e.g., different covariates, hyperparameters, or architectures).

\noindent
When many models are in use, maintaining them can be very
time-consuming and costly,
and, more importantly, a complete refitting of a pricing model often leads to
bigger changes in the feature contributions of the different risk factors due to the inherent correlation between the covariates in the training data and in smaller models due to statistical noise.
This can lead to significant changes in the pricing schemes of individual
policies, particularly, for segments that are not well-represented in the
training data.
Such changes are often undesirable from a business perspective, as they can
lead to unstable pricing over time and dissatisfaction among customers.

\noindent
For these reasons, model updates are often performed on a fixed time schedule,
such as every one or two years, rather than being based on performance metrics.
This, however, can lead to unnecessary updates and a suboptimal cost-benefit
ratio in the model update process, as time might have been better spent on
updating another model.
Moreover, performing updates on a fixed time schedule can result in slow
adaptation to changing market environments.

\noindent
This work concentrates on \textit{model monitoring}.
We establish theoretical foundations for key evaluation metrics, notably
deriving the asymptotic distribution of the sample Gini score.
By applying Murphy's score decomposition of the deviance loss, we assess global
and local auto-calibration.
Building on this, we introduce a framework for assessing and monitoring the
temporal robustness of non-life insurance pricing models to guide decisions on
model refitting.
For practical applicability, we present an illustrative example based on a
modified real-world dataset in which we inject controlled levels of concept
drift and discuss practical considerations and common pitfalls for real-world
implementations.

\bigskip

\noindent
\textbf{Organization of this manuscript}
In Sect.~\ref{sec:theoretical_background_related_work}, we outline the
theoretical background, review the relevant literature, discuss \textit{virtual
	drift} and \textit{real concept drift}, and develop the theoretical foundation
for our framework, including auto-calibration, evaluation metrics and their
properties, including the asymptotic behavior of the sample Gini score.
Sect.~\ref{sec:methodology} introduces the hypothesis testing framework for
\textit{model monitoring} over time, illustrates it using a modified real-world
insurance dataset, and discusses practical considerations and common pitfalls.
Finally, Sect.~\ref{sec:conclusion} concludes with a discussion and an outlook.

%% file: 01_Sec2_TechDetail_RelWork.tex
\section{Theoretical Background and Related Work}\label{sec:theoretical_background_related_work}
\noindent
This section establishes the foundational concepts, definitions, and notation
used throughout the paper.
As is common in regression modelling, we consider the random triplet
$(Y,\mathbf{X},V)$ on an underlying probability space, where $Y$ is a
non-negative real-valued response with finite mean, $\mathbf{X}$ denotes the
covariate vector and $V>0$ is a strictly positive exposure, a.s.
We denote the family of
potential distributions of $(Y, \mathbf{X},V)$ by $\mathcal{F}$, the conditional
distribution of $Y$, given $(\mathbf{X}, V)$, by $F_{Y \mid \mathbf{X}, V}$ and the
mean functional $T$ by
\begin{equation}\label{eq:true_regression_function}
	F_{Y \mid \mathbf{X}, V} \mapsto T(F_{Y \mid \mathbf{X}, V}) = \E[Y |
		\mathbf{X}, V] =\E[Y | \mathbf{X}] =:\mu(\mathbf{X}).
\end{equation}
Note that we adopt the common actuarial assumption that the conditional mean of
$Y$ is independent of the exposure $V$, i.e., $Y$ is an exposure-scaled
quantity, such as a claim frequency or a claim rate.

\noindent
The main goal in supervised learning is to estimate this true regression
function $\mu(\cdot)$ from a given dataset $\mathcal{L}={\{\left(y_{i},
	\mathbf{x}_{i}, v_{i}\right)\}}_{i=1}^{n}$ of i.i.d.~realizations of $(Y, \mathbf{X},V)$.
We denote the resulting estimator by $\hat{\mu}(\cdot)$; and $\mathcal{L}$ is
the learning dataset used to train this estimator, while we assume of holding a
second holdout dataset $\mathcal{T}$ that is going to be used for {\it model
		monitoring}.

%
%
%
\subsection{Virtual-Drift and Concept-Drift}\label{sec:virtual_concept_drift}
\noindent
The usual assumption in statistics is that we have an $\mathrm{i.i.d.
	}$ sample $(Y_i,\mathbf{X}_i, V_i)_{i=1}^n$ following
the same law as $(Y,\mathbf{X},V)$.
A common further assumption is that the data-generating process is stationary,
i.e., that the distribution of $(Y,\mathbf{X},V)$ does not change over time.
In actuarial modeling, one often trains a model on data from a given period
(e.g., years 2020-2023), and one assumes this data is under global trend
assumptions representative of a later period (e.g., 2025), i.e., when the model
will be used in production.
However, while this assumption is typically reasonable when comparing training
and holdout datasets drawn from the same period, it is frequently violated in
production datasets due to changes in the underlying population, in customer
behavior, the data-collection process, and last but not least, the real-world
environment.

\medskip
\noindent
\textbf{Related Work History}: Detecting changes in the underlying data-generating process has been studied
extensively for decades, see, e.g., \cite{schlimmer_beyond_1986, widmer_learning_1996},  and it
remains an active area of research.
Although work evolves in parallel across several communities, a key
consolidation is provided by the survey articles of \cite{gama_survey_2014} and
\cite{lu_learning_2019}, who systematize methods, clarify definitions (e.g.,
\textit{virtual} vs.
\textit{real concept drift}), and summarize the
state-of-the-art up to 2014 and 2019, respectively.
We adopt their terminology in what follows.

\noindent
More recent advances and surveys include: for scenarios where data labeling is
challenging, see, e.g., \cite{ackerman_automatically_2021}; for neural
networks, see, e.g., the survey of \cite{rabanser_failing_2019}; for data drift
detection in large-scale systems, see \cite{mallick_matchmaker_2022}; and more
recently for drift detection using deep neural networks and autoencoders, see
\cite{hu_concept_2025}.
For a formulation of drift as a distribution process and a survey of the
literature on unsupervised drift detection, we refer to
\cite{hinder_one_a_2024}.

\medskip
\noindent
\textbf{Terminology}: Because the terminology is more prevalent in the machine learning community (e.g., in online learning of classification tasks and in process mining), and less so in actuarial science,
we provide a brief overview of the terminology in an actuarial context.
As mentioned in \cite{gama_survey_2014}, the literature uses many different
terms to refer to changes in the data-generating process over time.
Common expressions include \textit{data drift}, \textit{covariate shift},
\textit{virtual shift}, \textit{temporary drift}, \textit{sampling shift},
\textit{feature change}, \textit{concept drift}, \textit{conditional change}
and \textit{real concept drift}.
Moreover, this terminology is not used consistently across the literature.

\noindent
For consistency, we follow the definitions of \cite{gama_survey_2014} and
distinguish two main types of drift: \textit{virtual drift} and
\textit{real concept drift}.
The former, \textit{virtual drift}, refers to changes in the population
distribution $F_{\mathbf{X},V}$ of the features $(\mathbf{X},V)$ and,
importantly, it occurs without changing the conditional distribution $F_{Y \mid
			\mathbf{X},V}$.
Thus, the true regression function $\mu(\mathbf{X})$ given in
\eqref{eq:true_regression_function} remains unchanged, only the portfolio
composition changes.
By contrast, \textit{real concept drift} refers to changes in the conditional
distribution $F_{Y \mid \mathbf{X},V}$ and, thus, in the true regression
function $\mu(\mathbf{X})$.
Notably, this change in the conditional distribution can occur with or without
a change in the population distribution $F_{\mathbf{X},V}$.
Table \ref{tab:virtual_concept_drift} summarizes the definitions and common
alternative names for \textit{virtual drift} and \textit{real concept drift}
used in the literature.
\begin{table}[h]
	\centering
	\caption{\textit{virtual drift} and the \textit{real concept drift}:}
	\label{tab:virtual_concept_drift}
	\begin{tabular}{rcc}
		\toprule
		              & \textit{virtual drift}                     & \textit{real concept drift}           \\
		\midrule
		Definition    & Changes in $F_{\mathbf{X},V}$              & Changes in $F_{Y \mid \mathbf{X}, V}$ \\
		              & without changing $F_{Y \mid \mathbf{X},V}$ &                                       \\
		\midrule
		Alternative   & data drift,                                & data drift,                           \\
		names         & virtual shift,                             & conditional change,                   \\
		in literature & temporary drift,                           & concept drift,                        \\
		              & sampling shift,                            &                                       \\
		              & feature change                             &                                       \\
		              & covariate shift                            &                                       \\
		\bottomrule
	\end{tabular}
\end{table}
\noindent
Furthermore, we refer to changes in the distribution of the features
$F_{\mathbf{X},V}$ as \textit{covariate drift}; note that this can result in
either \textit{virtual drift} or \textit{real concept drift}.

\medskip
\noindent
We acknowledge that detecting \textit{virtual drift} is very important in
the insurance industry for understanding how the portfolio evolves over time.
However, our main interest lies in changes in predictive performance over time
for a given portfolio, and in identifying when a model for the response should
be updated correspondingly.
Therefore, we restrict our attention to \textit{real concept drift} rather than
\textit{virtual drift}.
The phenomenon in which the predictive performance of a deployed model degrades
over time is also referred to as \textit{model drift}.

\noindent
While we now established the different drift terms,
it is important to note, that even if only a \textit{virtual drift} occurs,
i.e., no change in the true regression function $\mu(\mathbf{X})$,
there can still be an actual change in the performance of the estimator
of the regression function, i.e., the estimated model $\hat{\mu}(\mathbf{X})$.
This is because the model $\hat{\mu}(\mathbf{X})$ may not be trained on a
sufficiently rich dataset and thus may not generalize well.
If the exposure of new data increases in regions where the model performs
poorly, its performance will degrade.
While insufficient data coverage is an important issue, in the following
theoretical section we focus on \textit{real concept drift}, and assume that
the model is trained on a sufficiently rich dataset.

\medskip
\noindent
\textbf{Real Concept Drift Types}: There are typically 4 types of reasons
distinguished for \textit{real concept drift}
discussed in the literature.
These are \textit{sudden or abrupt drift}, \textit{gradual drift},
\textit{incremental drift}, and \textit{recurrent drift}; see
\cite{lu_learning_2019} for a comprehensive overview and visualization.
In this manuscript, we focus on the first three types of drift.

\medskip
\noindent
\textbf{Concept Drift Detection Method Types}:
There are several methods for detecting \textit{real concept drift} that can be broadly
categorized into the following four main families.
These are:
\begin{itemize}
	\item Data Distribution-Based Methods:
	      Typical examples of such methods involve computing distribution distance
	      measures such as the Kolmogorov-Smirnov statistic, Wasserstein metric, Kullback-Leibler divergence
	      and Jensen-Shannon metric between the old data and the new
	      data.
	      See, for example, Section 4.3 of \cite{hinder_one_a_2024}.
	\item Dimensionality Reduction-Based Methods:
	      Another common approach is to compare reconstruction errors obtained via PCA or
	      autoencoders.
	      Furthermore, domain classifier approaches are often used in this context, in
	      which one trains a classifier to distinguish between old and new data.
	      If the classifier performs well, it indicates a significant difference between
	      the two distributions, suggesting the presence of \textit{covariate drift}.
	      See, for example, \cite{rabanser_failing_2019}.
	\item Error Rate-Based Methods: These algorithms
	      are typically used for classification tasks and are designed to monitor a
	      predictive model's performance across time windows.
	      When a statistically significant change in the error rate is detected, a drift
	      alarm is triggered.
	      Influential examples include the Drift Detection Method (DDM) of
	      \cite{gama_learning_2004}, the Statistical Test of Equal Proportions (STEPD) of
	      \cite{nishida_detecting_2007}, and the Adaptive Windowing (ADWIN) of
	      \cite{bifet_learning_2007}.
	\item Multiple Hypothesis Methods:
	      These drift detection methods combine multiple different algorithms either in parallel
	      or in a hierarchical manner to detect drift.
	      See, for example, Section 3.2.3 in \cite{lu_learning_2019}.
\end{itemize}
\noindent
The approach in this manuscript follows the tradition of error rate-based
methods such as DDM and STEPD, but adapts them from classification to
regression in an insurance context.
Instead of traditional classification error measures, our framework proceeds in
two steps: (i) evaluating a regression model's ranking performance, and (ii)
testing global and local calibration.
For step (i), we derive the asymptotic properties of the Gini score, which is a
purely rank-based score, and we propose its use for assessing chances in risk
ranking.
For step (ii), calibration is assessed via Murphy's score decomposition of the
deviance loss in combination with isotonic regression to ensure
auto-calibration.
%
%
%
\subsection{Metrics and Auto-Calibration}\label{sec:metrics}
We start with the deviance loss and Murphy's score decomposition.
We continue with auto-calibration as the embracing concept of our monitoring
framework.
We then present the Gini score that underpins the risk ranking in the
monitoring procedure.
%
%
\subsubsection{Deviance Loss} \label{sec:deviance}
To ensure rigorous model validation, one should rely on strictly consistent
scoring functions; see \cite{gneiting_strictly_2007} and
\cite{gneiting_making_2011}.
Most regression frameworks used in practice are based on the exponential
dispersion family (EDF);
\cite{jorgensen_properties_1986,jorgensen_exponential_1987} and
\cite{nelder_generalized_1972}.
The EDF provides a unified parametrization for a large class of distributions,
such has the Gaussian, Poisson, gamma and Bernoulli distributions.
This unified parametrization is especially suited for maximum likelihood
estimation (MLE).
In particular, the MLE of the selected EDF is obtained by minimizing the
corresponding deviance loss of the selected EDF, and these deviance losses give
the strictly consistent scoring functions within the EDF framework.
This concept of deviance loss scoring has widely been adopted for \textit{model
	comparison} in the statistical and actuarial community.

\noindent
For auto-calibration testing in the monitoring framework we use a
weight-normalized deviance loss given by
\begin{equation}\label{definition normalized deviance}
	S(\mathbf{Y}, \hat{\boldsymbol{\mu}}, \mathbf{V}) = \frac{1}{\sum_{i=1}^{n} V_i}
	\sum_{i=1}^{n} \frac{V_i}{\varphi} \, d(Y_i, \hat{\mu}_i, V_i),
\end{equation}
where the prediction for response $\mathbf{Y}$ is represented as $\hat{\boldsymbol{\mu}} =
	(\hat{\mu}(\mathbf{X}_i))_{i=1}^n \in \mathbb{R}^n$, $\varphi>0$ is the given dispersion parameter and $d(Y_i,
	\hat{\mu}_i, V_i)$ denotes the unit deviance of the selected EDF, defined as the following
difference between the log-likelihoods
\begin{align*}
	d(Y, \hat{\mu}, V) & = 2\,\frac{\varphi}{V} \,\left(\log(f(Y; h(Y), V/\varphi)) - \log(f(Y; h(\hat{\mu}), V/\varphi))\right)                                                                                     \\
	                   & = 2 \, \begin{cases}
		                            Yh(Y) - \kappa(h(Y)) - Yh(\hat{\mu}) + \kappa(h(\hat{\mu}))                                                                   & \text{if } Y \in \mathbb{M},         \\
		                            \sup_{\tilde{\theta} \in \Theta} \left[Y\tilde{\theta} - \kappa(\tilde{\theta})\right] - Yh(\hat{\mu}) + \kappa(h(\hat{\mu})) & \text{if } Y \in \partial\mathbb{M},
	                            \end{cases}
\end{align*}
where $\kappa(\cdot):\Theta \to \mathbb{R}$ is the cumulant function on the effective domain $\Theta$, and $h=(\kappa')^{-1}$ is the canonical link of the selected EDF; we refer to \cite{wuthrich_statistical_2023} for an extended discussion.
Generally, the mean domain $\mathbb{M}=\kappa'(\Theta)$ of the selected EDF is
a (possibly infinite) interval, and if the response $Y$ is in the boundary
$\partial\mathbb{M}$ of the mean domain, the unit deviance is obtained by the
above limit consideration, see formula (4.8) in
\cite{wuthrich_statistical_2023}.
We provide the explicit forms of the gamma and Poisson deviance losses in
Appendix \ref{app:explicit_form_deviance_loss_gamma_poisson}.

\noindent
Although useful for \textit{model comparison}, the deviance loss in the above form is less
suitable for \textit{model monitoring}: it is rather sensitive to outliers, therefore it may
trigger false alarms, moreover, it lacks an absolute scale across datasets.
We therefore use the Gini score for the risk ranking monitoring and Murphy's
score decomposition of the weight-normalized deviance loss for level
calibration testing.
These are introduced next, we start with auto-calibration because this notion
is needed for both the Gini score and Murphy's score decomposition.
%
%

\subsubsection{Auto-Calibration} \label{sec:auto_calibration}
We start by introducing auto-calibration.

\medskip
\begin{definition}[Auto-Calibration]\label{def:auto_calibration}
	A random variable $Z$ is an auto-calibrated forecast of a random variable $Y$ if
	\begin{equation*}
		\E[Y \mid Z] = Z \quad \text{a.s.}
	\end{equation*}
	A regression function $\hat{\mu }(\cdot )$ is called auto-calibrated for $(Y,\mathbf{X})$ if
	\begin{equation}\label{eq:auto-calibration_regression}
		\hat{\mu }(\mathbf{X}) = \E[Y \mid \hat{\mu }(\mathbf{X})] \quad \text{a.s.}
	\end{equation}
\end{definition}
\noindent
In insurance pricing, auto-calibration is an important property of a regression
function, because it ensures that
each price cohort $\hat{\mu }(\mathbf{X})$ is on average self-financing.
That is, $\hat{\mu }(\mathbf{X})$ covers the cohort's expected claims, thus,
avoiding systematic cross-financing.
Another valuable implication of auto-calibration is that it ensures the
regression function $\hat{\mu }(\cdot )$ is (globally) unbiased at the
portfolio level, which is a minimal requirement for insurance pricing.

\noindent
Starting from any regression function $\hat{\mu }(\cdot)$, the following recalibration (rc) step gives an auto-calibrated regression function, see  \cite{wuthrich_isotonic_2024},
\begin{equation}\label{rc step}
	\hat{\mu }_{rc}(\mathbf{X}) = \E[Y \mid \hat{\mu }(\mathbf{X})];
\end{equation}
this is proved by the power property of conditional expectations.

\noindent
As discussed in  \cite{wuthrich_isotonic_2024}, an isotonic regression can be fitted to the observed sample
$(y_i,\hat{\mu}(\mathbf{x}_i),v_i)_{i=1}^n$, yielding a monotone step function that serves
as an empirically local recalibrated model for $\hat{\mu}_{rc}(\cdot)$.

\subsubsection{Murphy's score decomposition} \label{sec:murphy_decomposition}
Murphy's score decomposition (\cite{murphy_new_1973}) splits the score $S(\mathbf{Y}, \hat{\boldsymbol{\mu}}, \mathbf{V})$
into three components: uncertainty (UNC), discrimination (DSC), and
miscalibration (MCB), that is,
\begin{equation}
	S(\mathbf{Y}, \hat{\boldsymbol{\mu}}, \mathbf{V}) = \text{UNC}(\mathbf{Y}, \mathbf{V}) - \text{DSC}(\mathbf{Y}, \hat{\boldsymbol{\mu}}, \mathbf{V}) +
	\text{MCB}(\mathbf{Y}, \hat{\boldsymbol{\mu}}, \mathbf{V}),
\end{equation}
where the three components are defined as follows
\begin{align}
	\text{UNC}(\mathbf{Y}, \mathbf{V})                         & = S(\mathbf{Y}, \bar{\boldsymbol{\mu}}, \mathbf{V}),                                                                          \\
	\text{DSC}(\mathbf{Y}, \hat{\boldsymbol{\mu}}, \mathbf{V}) & = S(\mathbf{Y}, \bar{\boldsymbol{\mu}}, \mathbf{V}) - S(\mathbf{Y}, \hat{\boldsymbol{\mu}}_{rc}, \mathbf{V}),  \label{eq:DSC} \\
	\text{MCB}(\mathbf{Y}, \hat{\boldsymbol{\mu}}, \mathbf{V}) & = S(\mathbf{Y}, \hat{\boldsymbol{\mu}}, \mathbf{V}) - S(\mathbf{Y}, \hat{\boldsymbol{\mu}}_{rc}, \mathbf{V}), \label{eq:MCB}
\end{align}
where $\bar{\boldsymbol{\mu}}=(\bar{\mu}, \ldots, \bar{\mu})^\top \in \mathbb{R}^n$ simply contains the empirical mean $\bar{\mu}$ of the responses $\mathbf{Y}$ (ignoring covariates $\mathbf{X}$),
and $\hat{\boldsymbol{\mu}}_{rc} =
	(\hat{\mu}_{rc}(\mathbf{X}_i))_{i=1}^n  \in \mathbb{R}^n$ are the predictions of a recalibrated version of model $\hat{\mu}(\cdot)$, see \eqref{rc step}.

\medskip
\begin{remark}\label{rem:MCB_ge_0}\normalfont
	Working with strictly consistent scoring function implies that
	the expected values of \eqref{eq:DSC} and \eqref{eq:MCB}  are lower bounded by zero, because the recalibrated regression function $\widehat{\mu}_{rc}(\cdot)$, given in
	\eqref{rc step}, precisely minimizes the strictly consistent expected scores.
	These positive lower bounds do not automatically carry over to their empirical
	counterparts, when one uses an isotonic regression for the auto-calibration
	step, it only holds (approximately) if the risk ranking obtained by the
	estimated regression function $\hat{\mu}(\cdot)$ is (sufficiently) accurate.
	Intuitively, this is the case because strictly consistent scoring gives an
	unconstraint minimization problem, i.e., without a side constraint of
	preserving a giving ranking (as in isotonic regression), and the two solutions
	will align if the risk ranking used in the isotonic regression step is correct.
\end{remark}

\medskip
\noindent
Alternatively to the isotonic recalibration step, we can apply a basic balance correction of a model $\hat{\mu}(\cdot)$ via a GLM step.
Let $h$ be again the canonical link of the chosen EDF.
We define the basic balance corrected model as
\begin{equation}
	\hat{\mu}_{\text{bc}}(\mathbf{X}_i)=h^{-1}\!
	\left(\hat{\beta}_0+\hat{\beta}_1\,h(\hat{\mu}_i)\right),
\end{equation}
again, $\hat{\mu}_i=\hat{\mu}(\mathbf{X}_i)$ denotes the predicted value of the first regression model,
$\hat{\beta}_0 \in \mathbb{R}$ and $\hat{\beta}_1 \in \mathbb{R}$ are the
parameters of the GLM that are fitted on the sample ${\{\left(Y_{i},
			\hat{\mu}_{i}, V_{i}\right)\}}_{i=1}^{n}$ and estimated by MLE under the canonical link choice.
The choice of the canonical link ensures that the resulting model is globally
unbiased because it fulfills the balance property; see \cite{LindholmW}.
Because $\hat{\mu}_{\text{bc}}(\cdot)$ is an affine transformation of
$\hat{\mu}(\cdot)$ on the link scale, and the resulting predictions satisfy the
portfolio balance property, we can interpret it as a basic global level-shift
correction of the first regression model $\hat{\mu}(\cdot)$.

\noindent
Using this basic global balance correction, we can further decompose the empirical miscalibration
statistic $\text{MCB}(\mathbf{Y}, \hat{\boldsymbol{\mu}}, \mathbf{V})$ given by \eqref{eq:MCB} into two parts:
the global miscalibration statistic (GMCB) and the local miscalibration statistic
(LMCB):
\begin{equation} \label{eq:MCB_into_GMCB_LMCB}
	\text{MCB}(\mathbf{Y}, \hat{\boldsymbol{\mu}}, \mathbf{V})
	= \text{GMCB}(\mathbf{Y}, \hat{\boldsymbol{\mu}}, \mathbf{V})
	+ \text{LMCB}(\mathbf{Y}, \hat{\boldsymbol{\mu}}, \mathbf{V}),
\end{equation}
where define
\begin{align}
	\text{GMCB}(\mathbf{Y}, \hat{\boldsymbol{\mu}}, \mathbf{V}) & =
	S(\mathbf{Y}, \hat{\boldsymbol{\mu}}, \mathbf{V}) -
	S(\mathbf{Y}, \hat{\boldsymbol{\mu}}_{\text{bc}}, \mathbf{V}), \label{eq:GMCB} \\
	\text{LMCB}(\mathbf{Y}, \hat{\boldsymbol{\mu}}, \mathbf{V}) & =
	S(\mathbf{Y}, \hat{\boldsymbol{\mu}}_{\text{bc}}, \mathbf{V}) -
	S(\mathbf{Y}, \hat{\boldsymbol{\mu}}_{rc}, \mathbf{V}). \label{eq:LMCB}
\end{align}
Here, $\hat{\boldsymbol{\mu}}_{\text{bc}}$ are the predictions of the balance
corrected model $\hat{\mu}_{\text{bc}}(\cdot)$ and
$\hat{\boldsymbol{\mu}}_{rc}$ is an isotonic regression fitted to the sample
$(Y_i,\hat{\mu}_{\text{bc}}(\mathbf{X}_i),V_i)_{i=1}^n$.

\medskip
\begin{remark}\normalfont
	\begin{itemize}
		\item 	Since $(\hat{\beta}_0,\hat{\beta}_1)$ in $\hat{\mu}_{\text{bc}}(\cdot)$ are
		      obtained by minimizing the (weighted) deviance loss over these two parameters, we have
		      \begin{equation*}
			      \text{GMCB}(\mathbf{Y}, \hat{\boldsymbol{\mu}}, \mathbf{V}) = S(\mathbf{Y},\hat{\boldsymbol{\mu}},\mathbf{V})-S(\mathbf{Y},\hat{\boldsymbol{\mu}}_{\text{bc}},\mathbf{V})\ge 0,
		      \end{equation*}
		      since $(\hat{\beta}_0,\hat{\beta}_1)=(0,1)$ is a feasible minimization solution.
		\item Moreover, note that for $\hat{\beta}_1 > 0$ and strictly monotone and smooth functions $h(\cdot)$,
		      the ranking of $\hat{\mu}_{\text{bc}}(\cdot)$ is the same as that of $\hat{\mu}(\cdot)$,
		      because
		      \begin{equation*}
			      \hat{\mu}_1 < \hat{\mu}_2 \;\Longleftrightarrow\;  h^{-1}(\hat{\beta}_0+\hat{\beta}_1\,h(\hat{\mu}_1)) < h^{-1}(\hat{\beta}_0+\hat{\beta}_1\,h(\hat{\mu}_2))
		      \end{equation*}
		      and therefore,
		      \begin{equation*}
			      \hat{\mu}_1<\hat{\mu}_2 \;\Longleftrightarrow\;
			      \hat{\mu}_{\mathrm{bc}}(\hat{\mu}_1)<\hat{\mu}_{\mathrm{bc}}(\hat{\mu}_2).
		      \end{equation*}
		      Since the ranking is preserved by positive affine transformations on the link
		      scale, the isotonic recalibration $\hat{\mu}_{rc}(\cdot)$ calculated on the
		      balance-corrected model $\hat{\mu}_{\text{bc}}(\cdot)$ will also yield the same
		      result as calculated on the original model $\hat{\mu}(\cdot)$.
		      Note that by construction of the EDF, the canonical links $h$ are strictly
		      monotone and smooth, and $\hat{\beta}_1>0$ arises when $\hat{\mu}(\mathbf{X})$
		      is positively correlated with $Y$, which is typically satisfied in reasonable
		      regression models.
		\item Consequently, by the same reasoning as in Remark~\ref{rem:MCB_ge_0}, and since
		      the balance correction preserves the ordering whenever $\hat{\beta}_1>0$, we obtain under the canonical link choice the following:
		      if the risk ranking ability of $\hat{\mu}(\cdot)$ is sufficiently good, then we may
		      also expect, at the empirical level, that the local miscalibration component
		      satisfies
		      \begin{equation*}
			      \text{LMCB}(\mathbf{Y}, \hat{\boldsymbol{\mu}}, \mathbf{V})
			      = S(\mathbf{Y}, \hat{\boldsymbol{\mu}}_{\text{bc}}, \mathbf{V})
			      - S(\mathbf{Y}, \hat{\boldsymbol{\mu}}_{rc}, \mathbf{V}) \ge 0.
		      \end{equation*}
		      On the contrary, a negative LMCB indicates that the ranking ability of the
		      model is poor.
	\end{itemize}
\end{remark}
%
%

\medskip

\noindent
In our numerical example, we will test for auto-calibration, and we also are going to separate this into global and local miscalibration, as explained above.
We close this section with an auto-calibration test that is based on the
miscalibration statistic MCB.
Following the approach of \cite{delong_isotonic_2025}, which uses a parametric
bootstrap test based on the MCB statistic, we can test for the null hypothesis
that the regression model $\hat{\mu}(\cdot)$ is auto-calibrated.
This null hypothesis implies that the value of the MCB statistic in
\eqref{eq:MCB} is zero.
The test procedure is outlined in Algorithm~\ref{alg:autocalibration_test}.
For a detailed description of the variance estimation in Step~1 of
Algorithm~\ref{alg:autocalibration_test}, we refer to
\cite{delong_isotonic_2025}.
\begin{algorithm}[h!]
	\caption{MCB bootstrap auto-calibration test}\label{alg:autocalibration_test}
	\algrenewcommand\algorithmicrequire{\textbf{Input:}}
	\algrenewcommand\algorithmicensure{\textbf{Output:}}
	\begin{algorithmic}[1]
		\Require \begin{itemize}
			\item Holdout observations $\mathcal{T}=\{(y_i,\hat{\mu}_i, v_i)\}_{i=1}^{n}$ with $\hat{\mu}_i=\hat{\mu}(\mathbf{x}_i)$;
			\item Observed miscalibration statistic $\text{MCB}(\mathbf{y},\hat{\boldsymbol{\mu}},\mathbf{v})$ on $\mathcal{T}$;
			\item Assumed distribution family $F$ (parametrized by mean and variance) for $Y_i\mid \mathbf{x}_i$;
			\item Number of bootstrap replicates $B$;
			\item Significance level $\alpha$.
		\end{itemize}
		\Ensure p-value $p$ and
		decision on auto-calibration.
		\State \textbf{Variance estimation:}
		Estimate $\widehat{\mathrm{Var}}(Y_i\mid \mathbf{x}_i)$ by fitting an isotonic
		regression of the squared residuals on the predictions
		$\hat{\mu}(\mathbf{x}_i)$.
		\State \textbf{Bootstrap generation:}
		For $b=1,\dots,B$, sample independent responses $Y_i^{\star(b)} \sim
			F(\hat{\mu}(\mathbf{x}_i), \widehat{\mathrm{Var}}(Y_i\mid \mathbf{x}_i))$ and
		form $\mathcal{D}^{\star(b)}=\{(Y_i^{\star(b)},\hat{\mu}_i,v_i)\}_{i=1}^n$.
		\State \textbf{Isotonic recalibration:}
		Fit isotonic regressions on $\{(Y_i^{\star(b)},\hat{\mu}_i,v_i)\}_{i=1}^n$ to
		obtain $\hat{\mu}_{rc}^{(b)}(\mathbf{x}_i)$.
		\State \textbf{Bootstrap statistics:}
		Compute
		$\text{MCB}^{(b)}=\text{MCB}(\mathbf{Y}^{\star(b)},\hat{\boldsymbol{\mu}},\mathbf{v})$
		for $b=1,\dots,B$.
		\State \textbf{p-value:} $p=\frac{1}{B}\sum_{b=1}^B \1{\text{MCB}^{(b)} \ge \text{MCB}(\mathbf{y},\hat{\boldsymbol{\mu}},\mathbf{v})}$.
		\Statex \vspace{0.35ex}\hrule \vspace{0.35ex}
		\Statex \textbf{Decision rule:}
		Reject auto-calibration (null hypothesis) if $p<\alpha$.
		\Statex \vspace{0.1ex}\hrule
	\end{algorithmic}
\end{algorithm}
\noindent
%
%
\subsubsection{Gini score} \label{sec:gini_score}
While auto-calibration tests global and local level shifts, our monitoring
framework should also detect changes in risk rankings.
To capture such changes, we use the Gini score.
The Gini score is a rank-based metric that quantifies how well a model
discriminates between different responses.
It is a popular score in the machine learning community for \textit{model
	comparison}, especially, in a binary classification context.
It is less widely used in actuarial work, partly because the Gini score on its
own is not a strictly consistent scoring function for mean estimation;
optimizing it (i.e., maximizing it) does not necessarily lead to the best model
in regards to the true regression function $\mu(\mathbf{X})$, it only provides
the best risk ranking.
However, as has been shown in \cite{wuthrich_gini_2023}, on the class of
auto-calibrated regression functions, the Gini score is a suitable model
selection tool, as it selects the auto-calibrated model that has the
correct/best risk ranking.

\noindent
Before defining the Gini score, we note that there is not only one definition
in the literature.
For example, in \cite{denuit_testing_2024} the Gini score is defined purely via
the Lorenz curve, which depends only on the predictions and not on the response
variable $Y$.
This definition is popular and widely used in economics.
In \cite{frees_summarizing_2011} and \cite{frees_insurance_2014}, a Gini index
is defined via the so-called {\it ordered Lorenz curve}, which uses the
relativities, i.e., the ratios between premiums and scores, for the sorting.
In \cite{holvoet_neural_2025}, a version of this Gini index is used to
illustrate potential improvements in risk classification between two models.
In the machine-learning literature, the Gini score is often defined via the
Cumulative Accuracy Profile (CAP) curve.
An adaptation of this CAP-based definition to an actuarial context (also
accommodating ties in the risk ranking and case weights) is proposed in
\cite{brauer_wuethrich_gini_2025}.
In what follows, we adopt this latter version.
We first present the theoretical definitions of the CAP and the Gini score with
equal case weights $V=1$, and then introduce their empirical counterparts,
explicitly accounting for prediction ties and case weights $V>0$.

\medskip
\begin{definition}[Cumulative accuracy profile]\label{def:CAP}
	Let $\alpha \in (0,1)$.
	The CAP is defined as
	\begin{equation} \label{eq:CAP_definition}
		C_{Y, \hat{\mu}}(\alpha) =
		\frac{1}{\E[Y]} \E\left[Y \1{
				\hat{\mu} > F^{-1}_{\hat{\mu}}(1-\alpha)
			}\right] \in [0,1],
	\end{equation}
	where $F_{\hat{\mu }}$ is the distribution function of $\hat{\mu}(\mathbf{X})$ with left-continuous inverse $F^{-1}_{\hat{\mu }}$.
\end{definition}

\medskip

\noindent
Note that the concentration curve ($\CC$), which is more popular in the actuarial community
(see Definition 3.1 in \cite{denuit_model_2019}), is the mirrored version of the CAP, given by
$C_{Y, \hat{\mu}}(\alpha) = 1 - \CC_{Y, \hat{\mu}}(1 - \alpha)$.

\medskip
\noindent
\begin{definition}[Gini score]\label{def:gini_score}
	Using the CAP, the Gini score $G(Y, \hat{\mu})$ is defined as
	\begin{equation} \label{eq:gini_score}
		G(Y, \hat{\mu}) = \frac{\int_{0}^{1} C_{Y,\hat{\mu} } (\alpha ) \, d \alpha -
			\frac{1}{2} }{ \int_{0}^{1} C_{Y,Y } (\alpha ) \, d \alpha - \frac{1}{2}}.
	\end{equation}
\end{definition}

\begin{figure}[H]
	\centering
	\includegraphics[width=0.7\textwidth, trim={10pt 10pt 10pt 50pt}, clip]{
		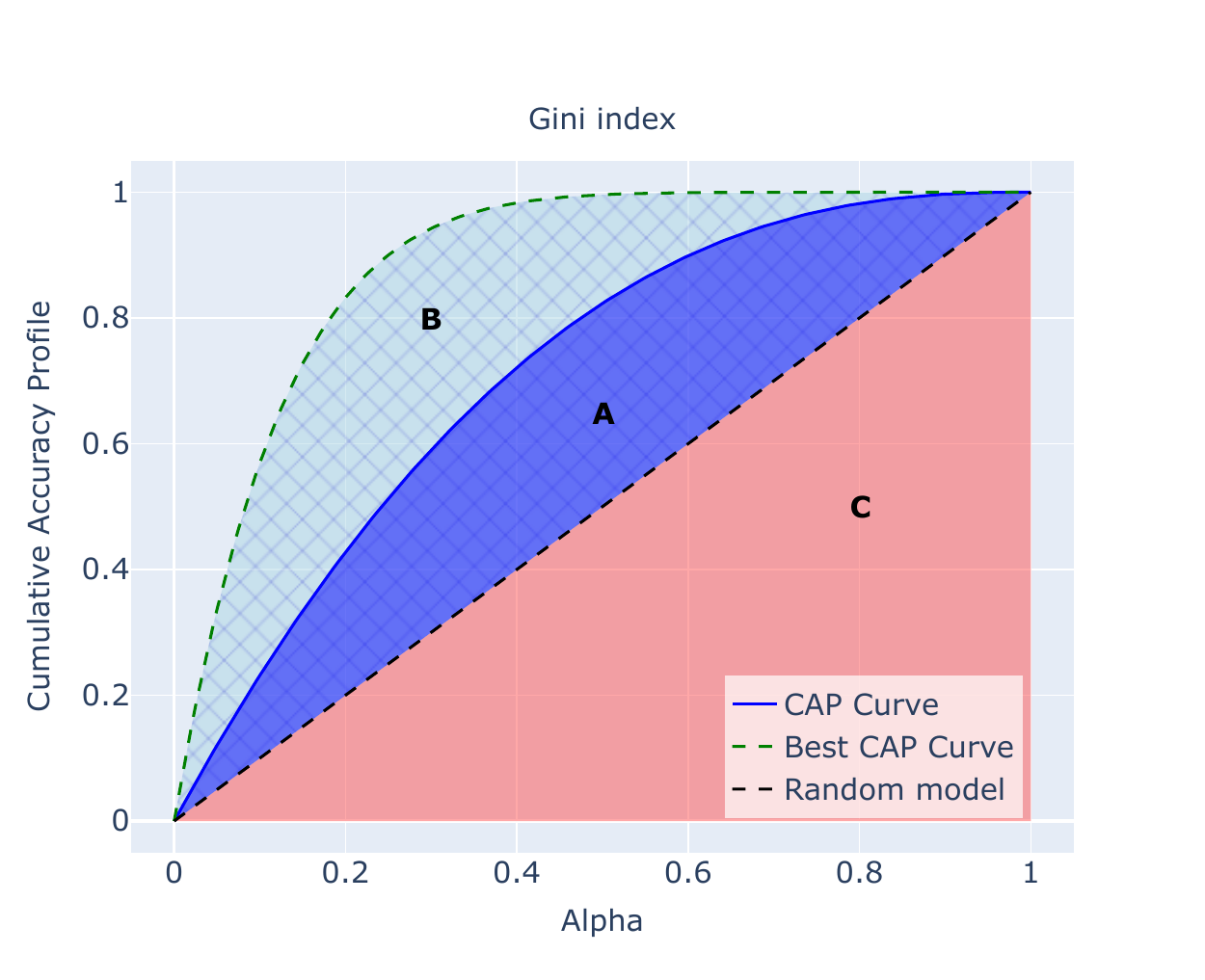}
	\caption{Geometric visualization of the Gini score.}\label{fig:geometric_gini_score}
\end{figure}

\noindent
Figure \ref{fig:geometric_gini_score} gives a geometric visualization of the involved CAP curves and the resulting Gini score, with areas $A$, $B$ and $C=1/2$ as illustrated in this figure, resulting in the Gini score
\begin{align}\label{eq:gini_geometric_interpretation}
	G(Y, \hat{\mu}) = \frac{\int_{0}^{1} C_{Y,\hat{\mu}} (\alpha ) d \alpha -
		\frac{1}{2} }{\int_{0}^{1} C_{Y,Y } (\alpha ) d \alpha - \frac{1}{2}} =
	\frac{(A + C) - C}{(B + C) - C} = \frac{A}{B}.
\end{align}
As the ranking induced by $\hat{\mu}$ becomes more consistent with the ordering
of $Y$, the CAP curve moves upward.
In the ideal case of perfectly aligned ranks, the CAP coincides with the best
CAP curve, that is $C_{Y,\hat{\mu}}(\alpha)=C_{Y,Y}(\alpha)$, for all
$\alpha\in(0,1)$, which implies $G(Y,\hat{\mu}) \le 1$.
We also notice that this is a purely rank-based measure, because if we select a
strictly increasing function $g$, then $G(Y, \hat{\mu})=G(Y, g(\hat{\mu}))$ as
this does not change the indicator event in \eqref{eq:CAP_definition}, or in
other words, the Gini score is invariant under strictly comonotonic
transformations of $\hat{\mu}$ as this does not change the ranking.

\bigskip
\noindent
Because we will use the Gini score in our monitoring framework, it is of
practical importance to understand the (asymptotic) behavior of its empirical version.
In the binary classification setting, the Gini score satisfies $G(Y,\hat{\mu})
	= 2\,\text{AUC}({Y,\hat{\mu}}) - 1$, where $\text{AUC}$ denotes the area under
the receiver operating characteristic (ROC) curve.
In this binary context, asymptotic normality of the empirical version of
$G(Y,\hat{\mu})$ follows from that of $\text{AUC}(Y,\hat{\mu})$; see
\cite{delong_comparing_1988}.
Furthermore, an asymptotic normality result for the economic Gini index (which
is different from the (machine learning) Gini score) is provided by Section 3
in \cite{davidson_reliable_2009}.
In addition, \cite{frees_summarizing_2011} provide asymptotic normality results
for a Gini index defined via the ordered Lorenz curve.

\medskip
\noindent
Assume there is a fixed regression function $\hat{\mu}(\cdot)$.
This gives us the predictor $\hat{\mu}(\mathbf{X})$ for $Y$.
In order to simplify the notation in the following theorem, we abbreviate
$\hat{\mu}:=\hat{\mu}(\mathbf{X})$ and $\hat{\mu}_i:=\hat{\mu}(\mathbf{X}_i)$,
so that we can interpret the predictors as real-valued random variables.
Moreover, for the resulting two-dimensional random vector we rewrite
$(Y,\hat{\mu}) \sim F_{Y,\hat{\mu}}$.
Intuitively, the bigger the (rank-)correlation within $F_{Y,\hat{\mu}}$, the
bigger the Gini score $G(Y, \hat{\mu})$.

\medskip
\noindent
\begin{theorem}[Asymptotic Normality of the machine-learning Gini score]\label{thm:gini_asymptotic_normality}
	Assume $(Y,\hat{\mu}) \sim F_{Y,\hat{\mu}}$ with finite first moments $\E[Y] <
		\infty$ and $\E[\hat{\mu}] < \infty$.
	Moreover, assume that the marginal distributions of $Y$ and $\hat{\mu}$ are
	continuous.
	Let $(Y_i,\hat{\mu}_i)$, $i \ge 1$, be i.i.d.~copies of $(Y,\hat{\mu})$.

	\noindent
	There exists a fixed variance parameter $\sigma^2 > 0$ such that we have asymptotic normality
	\begin{equation*}
		\sqrt{n}\left(\widehat{G}_n(Y,\hat{\mu}) - G(Y,\hat{\mu})\right)
		\xrightarrow{d}
		\mathcal{N}(0, \sigma^2).
	\end{equation*}
	where $\widehat{G}_n(Y,\hat{\mu})$ is the empirical (finite sample) Gini score as defined in Definition \ref{def:empirical_gini_score} below.
\end{theorem}

\medskip
\noindent
The proof is provided in Appendix \ref{app:proof_gini_asymptotic_normality}.

\medskip
\noindent
We note that we do not explicitly derive the asymptotic variance of the Gini
score described in Theorem \ref{thm:gini_asymptotic_normality}.
However, because we showed in the proof of Theorem
\ref{thm:gini_asymptotic_normality} that the functional $T$ is Hadamard
differentiable, we now know that the bootstrap approach is consistent; see
Theorem 3.21 in \cite{wasserman_all_2006}.
This is exploit in Algorithm \ref{alg:gini_bootstrap}.

\medskip

\begin{algorithm}[h!]
	\caption{Estimation of asymptotic normal parameters of the Gini score}\label{alg:gini_bootstrap}
	\algrenewcommand\algorithmicrequire{\textbf{Input:}}
	\algrenewcommand\algorithmicensure{\textbf{Output:}}
	\begin{algorithmic}[1]
		\Require \begin{itemize}
			\item Holdout observations $\mathcal{T}=\{(y_i,\hat{\mu}_i)\}_{i=1}^{n}$;
			\item Number of bootstrap replicates $B$;
			\item Optional: Assumed distribution $F$ for $Y_i\mid \hat{\mu}_i$ (mean/variance parametrization).
		\end{itemize}
		\Ensure Estimates $\widehat{\E}\!\left[G(Y, \hat{\mu})\right]$ and $\widehat{\sigma}\!\left[G(Y, \hat{\mu})\right]$.
		\State \textbf{Bootstrap resampling}:
		For each $b=1,\ldots,B$ draw $\mathcal{D}^{(b)}=\{(y_j,\hat{\mu}_j)^{(b)}\}_{j=1}^n$ by sampling $n$ instances with replacement from $\mathcal{T}$.
		\State \textbf{Optional: Bootstrap generation:}
		Estimate $\widehat{\mathrm{Var}}(Y\mid \hat{\mu}_i)$ by fitting an isotonic
		regression of the squared residuals on the predictions.
		For $b=1,\dots,B$ sample independent $Y_j^{\star(b)} \sim F(\hat{\mu}_j^{(b)},
			\widehat{\mathrm{Var}}(Y\mid \hat{\mu}_j^{(b)}))$ and form
		$\mathcal{D}^{\star(b)}=\{(Y_j^{\star},\hat{\mu}_j)^{(b)}\}_{j=1}^n$.
		\State \textbf{Compute bootstrap Gini scores}:
		For each $b$, compute the empirical Gini score $\widehat{G}^{(b)} = \widehat{G}_n(Y, \hat{\mu})$ on $\mathcal{D}^{(b)}$ (optionaly on $\mathcal{D}^{\star(b)}$).
		\State \textbf{Aggregate}:
		\begin{equation*}
			\widehat{\E}\!\left[G(Y, \hat{\mu})\right]     =\frac{1}{B}\sum_{b=1}^B \widehat{G}^{(b)}
			\quad and \quad
			\widehat{\sigma}\!\left[G(Y, \hat{\mu})\right] =\sqrt{\frac{1}{B-1}\sum_{b=1}^B \left(\widehat{G}^{(b)}-\widehat{\E}\!\left[G(Y, \hat{\mu})\right]\right)^2}.
		\end{equation*}
	\end{algorithmic}
\end{algorithm}

\medskip
\noindent
Algorithm \ref{alg:gini_bootstrap} offers two distinct bootstrap strategies.
Step 1 implements a \textit{random design} approach by resampling the
observations.
This non-parametric method is generally preferred as it remains valid
regardless of the underlying distribution of the features.
The optional Step 2 represents a parametric bootstrap approach.
Here, the predictions $\hat{\mu}_i$ are used to generate new responses from an
estimated conditional distribution.
This alternative is appropriate when one is confident in the distributional
assumptions of, e.g., a GLM.
In this case, Step~2 may yield more accurate estimates of $\widehat{\E}\!
	\left[G(Y, \hat{\mu})\right]$ and $\widehat{\sigma}\!\left[G(Y, \hat{\mu})\right]$
in small samples by exploiting the assumed true data-generating mechanism.

\begin{figure}[h]
	\centering
	\begin{subfigure}{0.49\textwidth}
		\centering
		\includegraphics[width=\textwidth, trim={20pt 25pt 55pt 55pt}, clip]{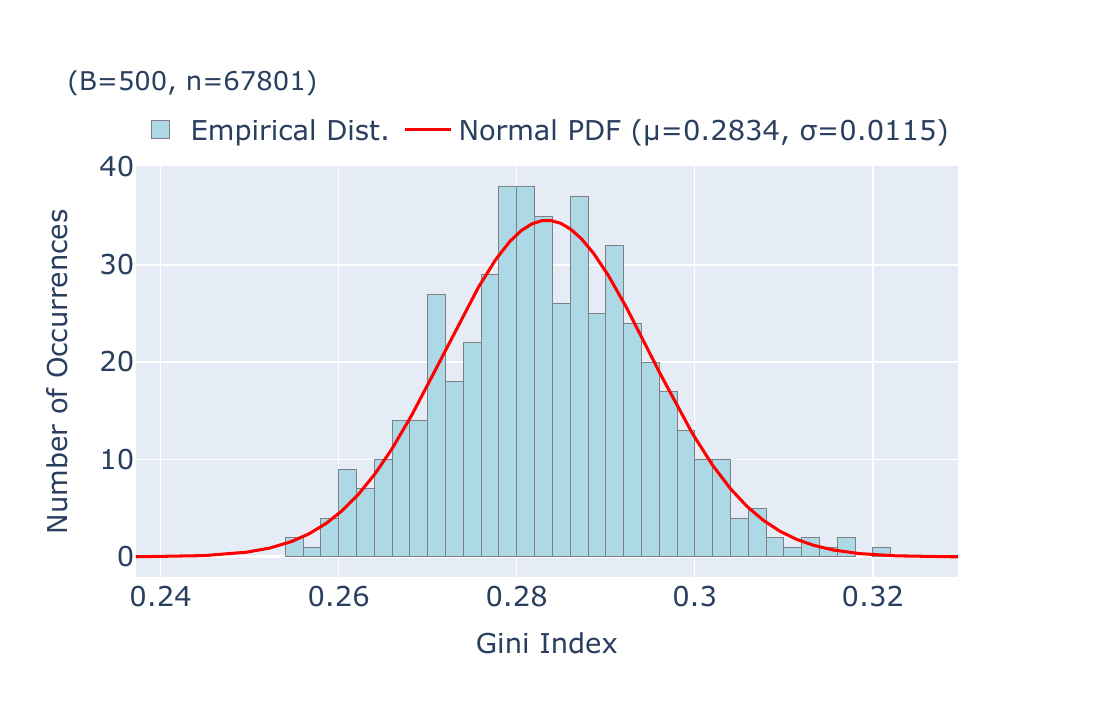}
		\caption{$B=500$ and $n=67801$}
	\end{subfigure}
	\hfill
	\begin{subfigure}{0.49\textwidth}
		\centering
		\includegraphics[width=\textwidth, trim={20pt 25pt 55pt 55pt}, clip]{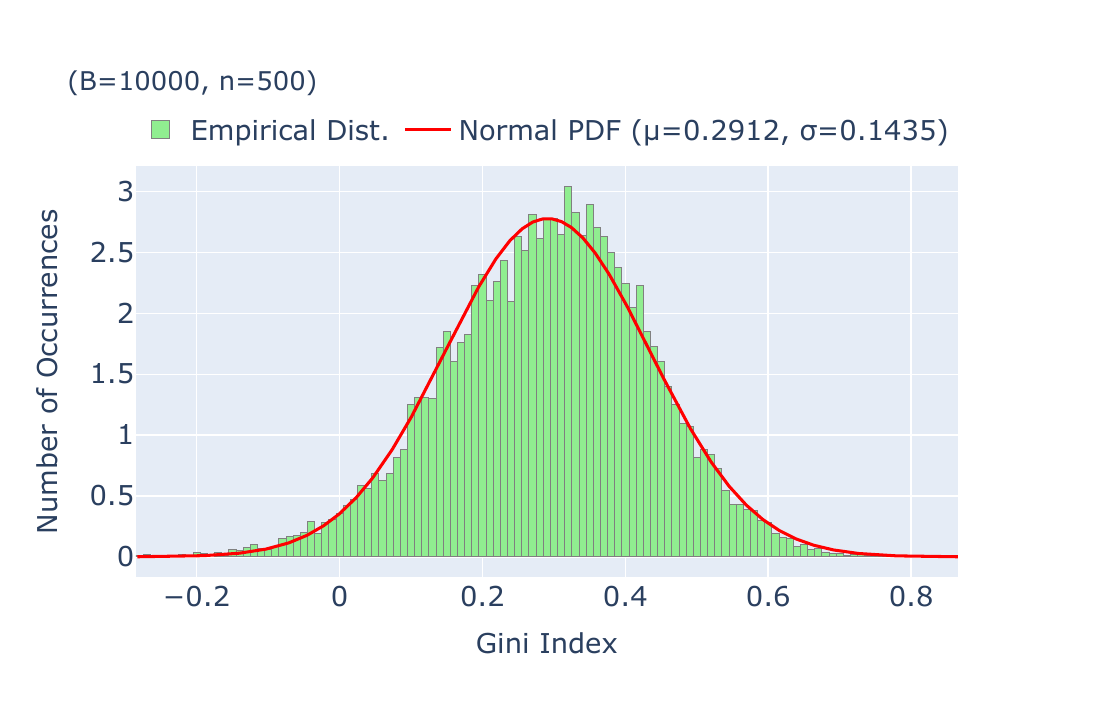}
		\caption{$B=10000$ and $n=500$}
	\end{subfigure}
	\begin{subfigure}{0.49\textwidth}
		\centering
		\includegraphics[width=\textwidth, trim={20pt 25pt 55pt 55pt}, clip]{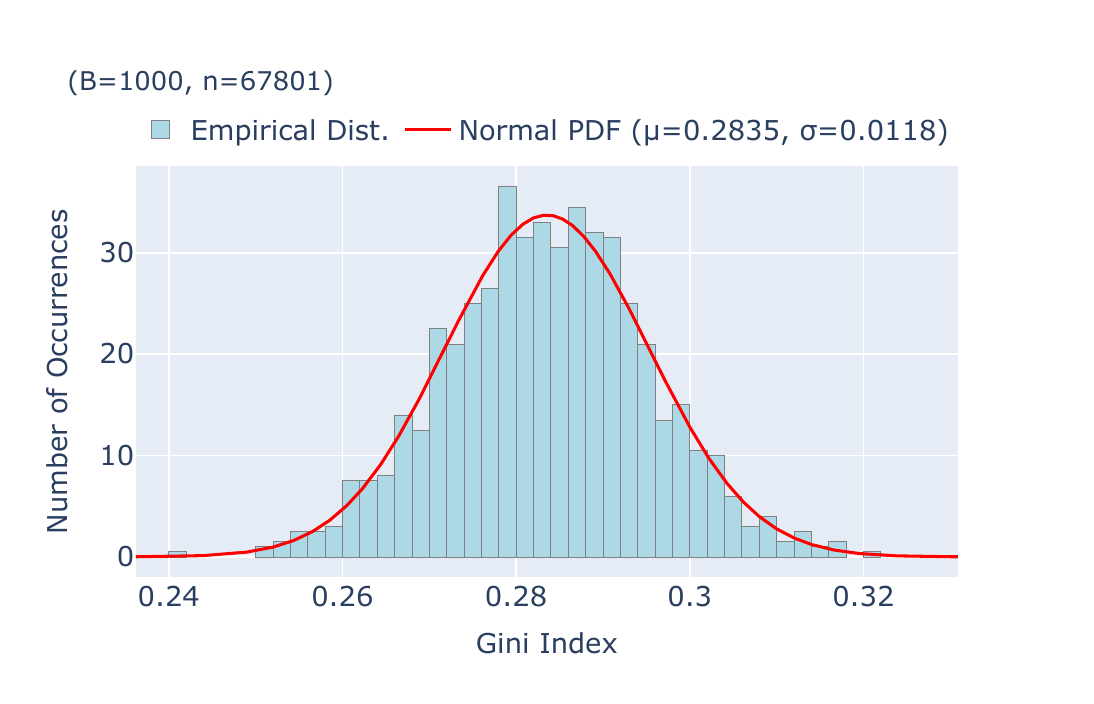}
		\caption{$B=1000$ and $n=67801$}
	\end{subfigure}
	\hfill
	\begin{subfigure}{0.49\textwidth}
		\centering
		\includegraphics[width=\textwidth, trim={20pt 25pt 55pt 55pt}, clip]{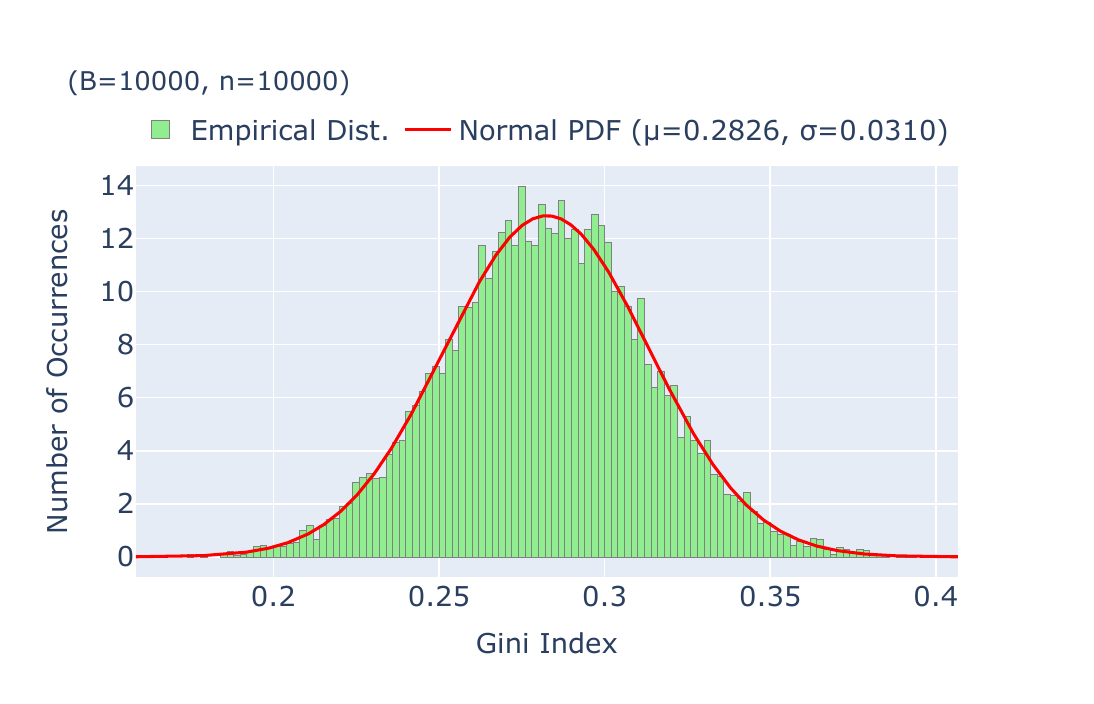}
		\caption{$B=10000$ and $n=10000$}
	\end{subfigure}
	\begin{subfigure}{0.49\textwidth}
		\centering
		\includegraphics[width=\textwidth, trim={20pt 25pt 55pt 55pt}, clip]{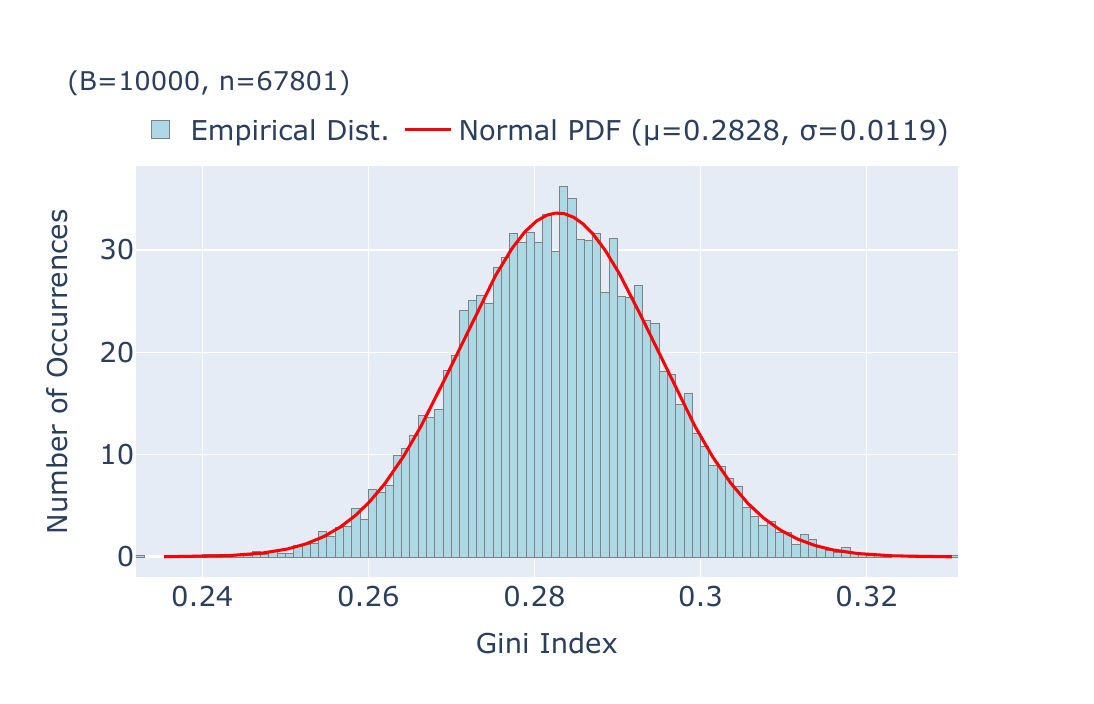}
		\caption{$B=10000$ and $n=67801$}
	\end{subfigure}
	\hfill
	\begin{subfigure}{0.49\textwidth}
		\centering
		\includegraphics[width=\textwidth, trim={20pt 25pt 55pt 55pt}, clip]{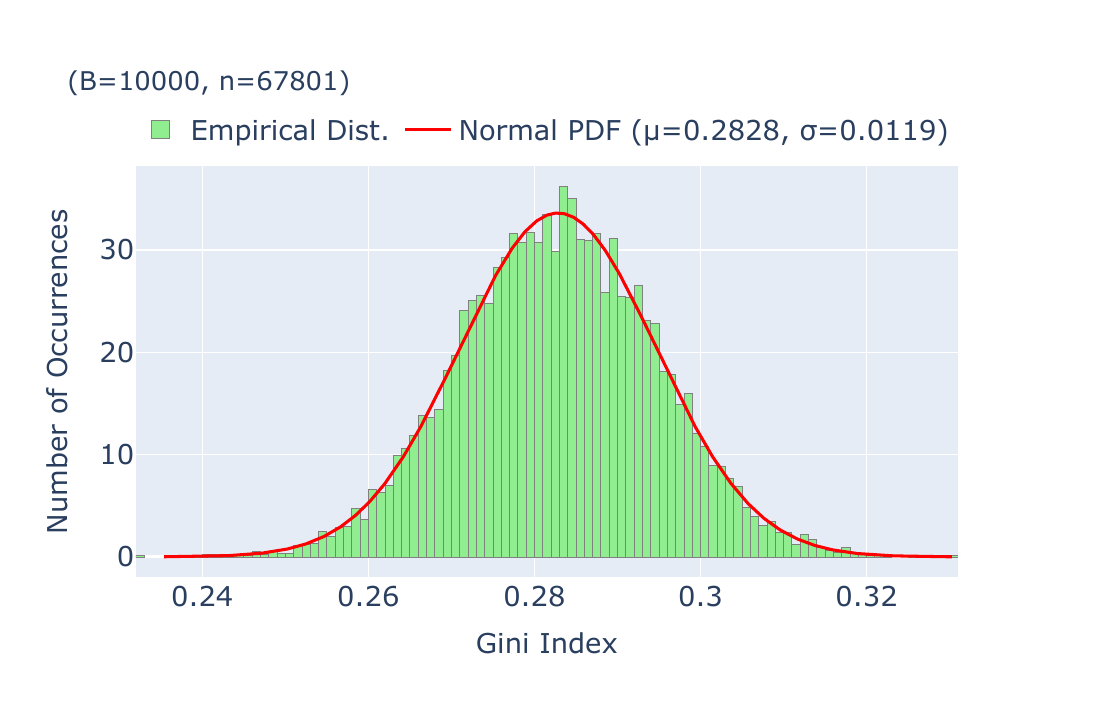}
		\caption{$B=10000$ and $n=67801$}
	\end{subfigure}
	\caption{Histograms of the bootstrap Gini indices for varying numbers of bootstrap samples $B$ (left) and varying holdout sample sizes $n$ (right).}
	\label{fig:bootstrap_gini_freMTPL}
\end{figure}

\medskip
\noindent
In Figure~\ref{fig:bootstrap_gini_freMTPL}, we provide a visual illustration of the (asymptotic) normality of the Gini score.
We implement Algorithm~\ref{alg:gini_bootstrap} for the dataset and model
described in Section~\ref{subsec:illustration_example}, varying the number of
bootstrap replicates $B$ (left side) and the holdout sample size $n$ (right
side).
As expected, the empirical distributions of the bootstrap Gini indices are well
approximated by a normal distribution, and this approximation improves as the
number of bootstrap replicates $B$ increases.
Moreover, the empirical standard deviation of the Gini score decreases with
larger holdout sample sizes $n$.
This behavior is consistent with the intended use of our \textit{model
	monitoring} framework: when the holdout sample size is small, the underlying
model has typically been trained on limited data, so the resulting Gini score
is more variable and deviations from the reference value are harder to detect
and reject.

\medskip
\noindent
We still need to adapt the above empirical version of the Gini score to actuarial practice.
First, in actuarial practice, it is likely to have ties in the predictors
$\hat{\mu}_i$, e.g., as soon as we have two policyholders with identical
covariates $\mathbf{X}$, we obtain a tie in this covariate and predictor,
respectively.
Therefore, in practice, we cannot generally assume that the marginal
distributions are continuous -- of course, this problem originates from the
fact that all continuous variables are measured with finite precision, e.g.,
the age of the policyholder is recorded in yearly units.
Secondly, we still need to adapt the Gini score to case weights $V>0$.
This is done as described in \cite{brauer_wuethrich_gini_2025}.

\medskip

\begin{definition}[Empirical Gini score]\label{def:empirical_gini_score}
	The empirical Gini score is defined as follows
	\begin{equation}\label{eq:Gini score empirical}
		\widehat{G}(\mathbf{y}, \hat{\boldsymbol{\mu}}, \mathbf{v}) ~=~ \frac{(A^\downarrow+A^\uparrow)/2}{B} ~\le~ 1,
	\end{equation}
	where $A^\downarrow$, $A^\uparrow$ and $B$ are given below.
\end{definition}

\medskip
\noindent
$B$ is an empirical estimate of the area between the best CAP curve and
the diagonal; see \eqref{eq:gini_geometric_interpretation}.
Based on observed responses $\mathbf{y}$, we first construct the order
statistics $y_{(1)} \ge \ldots \ge y_{(n)}$.
This gives us a ranking (illustrated by the round brackets), and we map this
ranking to the observed case weights $\mathbf{v}$ giving the ordered sample
$(v_{[i]})_{i=1}^n$ (square brackets indicate an implied ranking, in this case
from the responses).
For $0 \le i \le n$ we then set
\begin{equation*}
	\alpha_i  = \frac{1}{ \sum_{j=1}^n v_j}\, \sum_{j=1}^i v_{[j]}
	\qquad \text{and} \qquad
	\widehat{L}_n(\alpha_i)
	=\frac{1}{\sum_{j=1}^n v_j y_j} \,
	\sum_{j=1}^i  v_{[j]}
	y_{(j)}.
\end{equation*}
The latter is an empirical version of the mirrored Lorenz curve, see
\cite{brauer_wuethrich_gini_2025}.
With this notation, $B$ is defined by
\begin{equation*}\label{eq:B_estimate}
	B = \sum_{i=1}^n \frac{\widehat{L}_n(\alpha_i) +
		\widehat{L}_n(\alpha_{i-1})}{2}\, \left(\alpha_i - \alpha_{i-1}\right) -
	\frac{1}{2}.
\end{equation*}
Concerning the numerator in \eqref{eq:Gini score empirical}, it is given by the
average of the two areas $A^\downarrow$ and $A^\uparrow$, which provide two
empirical estimates of the area between the CAP curve and the diagonal; see
\eqref{eq:gini_geometric_interpretation}.
We proceed as follows.
We first order the predictions in decreasing order, $\hat{\mu}_{(1)} \ge \dots
	\ge \hat{\mu}_{(n)}$, and, in the presence of ties, we consider the decreasing
and increasing suborders induced by the responses $\mathbf{y}$, respectively,
in these ties.
These two (sub-)orders are then mapped to the responses (with indexed in square
brackets) $(y_{[i\downarrow]})_{i=1}^n$ and $(y_{[i\uparrow]})_{i=1}^n$ and the
case weights $(v_{[i\downarrow]})_{i=1}^n$ and $(v_{[i\uparrow]})_{i=1}^n$.
Thus, we have two versions, from the two different sub-orders in the ties of
$\hat{\boldsymbol{\mu}}$.
Using these ordered triples, we construct the empirical CAP curves for
$0 \le i \le n$ by
\begin{equation*}
	\alpha^\downarrow_i  = \frac{1}{ \sum_{j=1}^n v_j}\,\sum_{j=1}^i v_{[j\downarrow]}
	\qquad \text{and} \qquad
	\widehat{C}^\downarrow_n(\alpha^\downarrow_i)
	=\frac{1}{\sum_{j=1}^n v_j y_j} \,
	\sum_{j=1}^i  v_{[j\downarrow]}
	y_{[j\downarrow]}.
\end{equation*}
and define the associated area $A^\downarrow$ as
\begin{equation*}
	A^\downarrow = \sum_{i=1}^n \frac{\widehat{C}^\downarrow_n(\alpha^\downarrow_i)
		+ \widehat{C}^\downarrow_n(\alpha^\downarrow_{i-1})}{2}\,
	\left(\alpha^\downarrow_i - \alpha^\downarrow_{i-1}\right) - \frac{1}{2}.
\end{equation*}
Analogously, we define the empirical CAP curve and the corresponding area for
the worst sub-order induced by the responses, yielding $A^\uparrow$.
Further details can be found in \cite{brauer_wuethrich_gini_2025}.
In particular, the computation of the empirical Gini score -- though a bit
technical here -- is straightforward, and \cite{brauer_wuethrich_gini_2025}
give a short computer code.

%% file: 01_Sec3_Framework_for_Model_Monitoring.tex
\section{Model Monitoring Framework}
\label{sec:methodology}
First, we outline our general framework for \textit{model monitoring}, then we
provide an example for illustration, and we close by discussing common pitfalls
and practical considerations.
%
%
%
\subsection{General Framework for Model Monitoring}
\label{subsec:general_framework_model_monitoring}
\noindent
On the one hand, it is important not to miss a necessary update of a model in
order to maintain the model's performance.
But on the other hand, as noted in the introduction, the model update process
is time-consuming, complex, and may introduce instability into the pricing
model.
Therefore, changes should not be made lightly.
This, in turn, prompts the question of whether an update is necessary.

\noindent
We first outline the model update process to clarify which information is
available at the decision point.
We do this by providing an explicit example of an annual model update process
with a fixed window size, though the same logic applies to other update
frequencies.
In a typical annual cycle, the incumbent model is either recalibrated with new
data or replaced by a newly developed model that incorporates the latest data.
Because models are trained on prior-year data while data from the update year
is not yet fully observed or validated, a time lag arises.
This time lag is further increased by the time required for model development,
validation and governance.
Therefore, in an annual cycle, a one-year time lag can arise.
Consequently, the most recent data used to train or assess the model comes from
the year preceding the update.

\begin{figure}[h!]
	\centering
	\begin{tikzpicture}
		\draw[thick] (0,0) -- (13,0);
		\foreach \x/\desc in {0.5, 2, 3.5, 5, 6.5, 9.5, 12.5} {
				\draw[thick] (\x,0.2) -- (\x,-0.2);
			}
		\foreach \x/\desc in {1.25/2020, 2.75/2021, 4.25/2022, 5.75/2023, 8/2024, 11/2025} {
				\node[below] at (\x,-0.2) {\desc};
			}
		\draw[decorate,decoration={brace,mirror,amplitude=10pt}] (0.6,-0.5) -- (6.4,-0.5)
		node[midway,below,yshift=-10pt] {training data};

		\draw[decorate,decoration={brace,mirror,amplitude=10pt}] (6.6,-0.5) -- (9.4,-0.5)
		node[midway,below,yshift=-10pt,align=left] {$\hat{\mu}_{2024}$ development \\ and deployment};

		\draw[decorate,decoration={brace,mirror,amplitude=10pt}] (9.6,-0.5) -- (12.4,-0.5)
		node[midway,below,yshift=-10pt] {prediction period};
	\end{tikzpicture}
	\caption{Example timeline of a model update cycle in 2024.
		The index $2024$ in $\hat{\mu}_{2024}$ indicates the year in which the model is
		developed.
	}
	\label{fig:example_timeline_model_update_2024}
\end{figure}
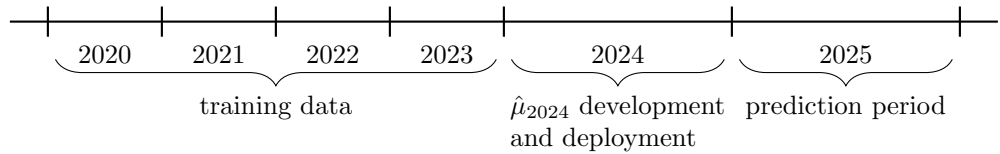
\noindent
In Figure \ref{fig:example_timeline_model_update_2024}, an example timeline of
a model update cycle in 2024 is shown.
In this example, claim data from calendar years 2020--2023 is used to train a
model.
The model $\hat{\mu}_{2024}(\cdot)$ is then developed in 2024 without using the
2024 data for training or validation, i.e., this model is assumed to be tested
and calibrated in 2024 using data from 2020--2023.
This model is subsequently deployed to predict and price, e.g., claim counts
for 2025.

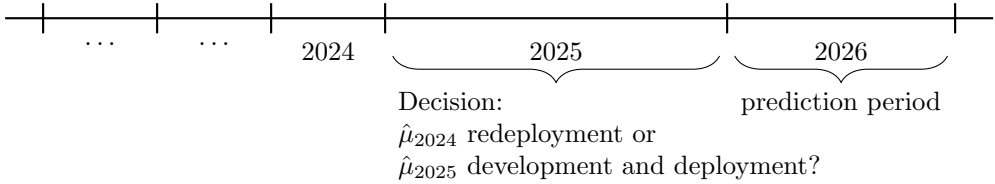
\begin{figure}[h!]
	\centering
	\begin{tikzpicture}
		\draw[thick] (0,0) -- (13,0);
		\foreach \x/\desc in {0.5, 2, 3.5, 5, 9.5, 12.5} {
				\draw[thick] (\x,0.2) -- (\x,-0.2);
			}
		\foreach \x/\desc in {1.25/$\ldots$, 2.75/$\ldots$, 4.25/2024, 7.25/2025, 11/2026} {
				\node[below] at (\x,-0.2) {\desc};
			}
		\draw[decorate,decoration={brace,mirror,amplitude=10pt}] (5.1,-0.5) -- (9.4,-0.5)
		node[midway,below,yshift=-10pt,align=left] {\qquad \qquad Decision:
			\\
			\qquad \qquad $\hat{\mu}_{2024}$ redeployment or
			\\
			\qquad \qquad $\hat{\mu}_{2025}$ development and deployment?};

		\draw[decorate,decoration={brace,mirror,amplitude=10pt}] (9.6,-0.5) -- (12.4,-0.5)
		node[midway,below,yshift=-10pt] {prediction period};
	\end{tikzpicture}
	\caption{Example timeline for decision-making in 2025.
		The index $2024$ in $\hat{\mu}_{2024}$ indicates the year in which the model is
		developed.
	}
	\label{fig:example_timeline_model_update_2025}
\end{figure}

\noindent
The question we seek to answer in the 2025 update cycle is whether the model
created in 2024, $\hat{\mu}_{2024}$, using data from calendar years 2020--2023 can be reused
(with minor global level adjustments, e.g., for inflation) and deployed in 2025 to produce predictions and prices for
2026, or whether a new model, $\hat{\mu}_{2025}$, should be developed (being based on the data up to calendar year 2024).
An illustration of this timeline is shown in Figure
\ref{fig:example_timeline_model_update_2025}.
This decision should be made in 2025, when data for 2024 becomes available.

\medskip
\noindent
To guide this decision we apply a two-step monitoring framework:
\begin{enumerate}
	\item \textbf{Risk ranking monitoring (Gini test).}
	      Detect potential rank shifts by testing the discriminatory performance of
	      $\hat{\mu}_{2024}$ on the new 2024 data.
	      We use the Gini score with its derived asymptotic properties to form a
	      hypothesis test that signals deterioration in risk ranking performance.
	\item \textbf{Auto-calibration test (global and local).
	      }
	      Conditional on an acceptable ranking, assess calibration.
	      (a) Test global calibration via Murphy's score decomposition and a basic
	      balance-correction.
	      (b) Further, test local (cohort-level) calibration using isotonic regression-based
	      model calibration to identify if local level shifts exist.
\end{enumerate}
We illustrate the process for the Gini score test in
Algorithm~\ref{alg:gini_ranking_test}.

\begin{algorithm}[h!]
	\caption{Gini-based ranking drift test}\label{alg:gini_ranking_test}
	\algrenewcommand\algorithmicrequire{\textbf{Input:}}
	\algrenewcommand\algorithmicensure{\textbf{Output:}}
	\begin{algorithmic}[1]
		\Require \begin{itemize}
			\item Current model $\hat{\mu}_{old}(\cdot)$ (e.g., from train-period 2020--2023 with $\hat{\mu}_{old} = \hat{\mu}_{2024}$);
			\item Holdout data $\mathcal{T}_{old}=\{(y_i,\hat{\mu}_i, v_i)^{old}\}_{i=1}^{n_{old}}$ with $\hat{\mu}_i=\hat{\mu}_{old}(\mathbf{x}_i^{old})$;
			\item New-period data $\mathcal{T}_{new}=\{(y_j,\hat{\mu}_j, v_j)^{new}\}_{j=1}^{n_{new}}$ with $\hat{\mu}_j=\hat{\mu}_{old}(\mathbf{x}_j^{new})$;
			\item Significance level $\alpha$.
		\end{itemize}
		\Ensure Test statistic $z$, p-value $p$, and decision on ranking drift.
		\State \textbf{Estimate reference distribution (old period):}
		Using Algorithm~\ref{alg:gini_bootstrap} on $\mathcal{T}_{old}$, compute
		\vspace{-0.1cm}
		\begin{equation*}
			\widehat{\E}\!\left[G(Y, \hat{\mu}_{old})\right]
			\text{ and }
			\widehat{\sigma}\!\left[G(Y, \hat{\mu}_{old})\right].
		\end{equation*}
		\State \textbf{Compute new-period Gini score:}
		On a new-period sample $\mathcal{T}_{new}$ of comparable size and covariate distribution (choose $n_{new}\approx n_{old}$), compute
		\vspace{-0.1cm}
		\begin{equation*}
			\widehat{G}^{new} = \widehat{G}(\mathbf{y}^{new}, \hat{\boldsymbol{\mu}}_{old}(\mathbf{x}^{new}), \mathbf{v}^{new}).
		\end{equation*}
		\Statex \vspace{0.35ex}\hrule \vspace{0.35ex}
		\Statex \textbf{Null hypothesis (no real concept drift):}
		Under the assumption of no \textit{real concept drift}, the risk ranking
		remains the same.
		Therefore, the Gini score on the new data should come from the same distribution as the holdout Gini values from the training period, i.e., under the null hypothesis $H_0$:
		\vspace{-0.1cm}
		\begin{equation*}
			\widehat{G}^{new}
			\sim \mathcal{N}\!\Big(
			\widehat{\E}\!\left[G(Y, \hat{\mu}_{old})\right],\;
			\widehat{\sigma}\!\left[G(Y, \hat{\mu}_{old})\right]^2
			\Big).
		\end{equation*}
		\Statex \vspace{0.1ex}\hrule \vspace{0.35ex}
		\State \textbf{Compute test statistic:}
		\vspace{-0.1cm}
		\begin{equation*}
			z=\dfrac{\widehat{G}^{new}-\widehat{\E}\!\left[G(Y, \hat{\mu}_{old})\right]}{\widehat{\sigma}\!\left[G(Y, \hat{\mu}_{old})\right]}.
		\end{equation*}
		\State \textbf{p-value (two-sided):}
		$p=2\bigl(1-\Phi(|z|)\bigr)$, where $\Phi(\cdot)$ is the standard normal cdf.
		\Statex \vspace{0.35ex}\hrule \vspace{0.35ex}
		\Statex \textbf{Decision rule:}
		Reject $H_0$ if $p<\alpha$.
	\end{algorithmic}
\end{algorithm}

\medskip

\noindent
Regarding interpretation, the test statistic $z$ measures changes in Gini score
performance in standard-deviation units; negative values ($z<0$) indicate
deterioration and positive values indicate improvement.
For example, a $z$-value of $-1$ (corresponding to a p\text{-}value of $\approx
	0.32$) means that the Gini performance on the new data is one standard
deviation worse than the average Gini performance in the training period.

\medskip
\begin{remark}\normalfont
	Algorithm~\ref{alg:gini_ranking_test} is a two-sided test, as we want to detect
	both deterioration and improvement in model ranking performance, one can also
	use a one-sided test if only deterioration is of interest.
\end{remark}

\medskip

\noindent
Second, we assess auto-calibration at the global and local levels.
If auto-calibration is violated, this indicates that price levels have shifted
within price cohorts, implying that a model update is necessary.
	{
		\algrenewcommand{\alglinenumber}[1]{
			\ifnum#1=1 \footnotesize(a):\else
				\ifnum#1=2 \footnotesize(b):\else
					(#1)\fi\fi}
		\begin{algorithm}[h!]
			\caption{Auto-calibration drift test (global and local)}
			\label{alg:autocalibration_two_step}
			\algrenewcommand\algorithmicrequire{\textbf{Input:}}
			\algrenewcommand\algorithmicensure{\textbf{Output:}}
			\begin{algorithmic}[1]
				\Require \begin{itemize}
					\item Current model $\hat{\mu}_{old}(\cdot)$;
					\item New-period data $\mathcal{T}_{new}=\{(y_j,\hat{\mu}_j, v_j)^{new}\}_{j=1}^{n_{new}}$ with $\hat{\mu}_j=\hat{\mu}_{old}(\mathbf{x}_j^{new})$;
					\item Significance levels $\alpha_{global}$ and $\alpha_{local}$.
				\end{itemize}
				\Ensure p-values $p_{global}$ and $p_{local}$; decisions on global and local level shifts.
				\State \textbf{Global level shift test (GMCB).}
				Apply a modified version of Algorithm~\ref{alg:autocalibration_test} to
				$\mathcal{T}_{new}$ that replaces the isotonic recalibration
				$\hat{\mu}_{rc}(\cdot)$ in Step~3 with the balance-correction
				$\hat{\mu}_{\text{bc}}(\cdot)$ computed on $\mathcal{T}_{new}$, and uses the
				global component GMCB (\ref{eq:GMCB}) of the full miscalibration statistic MCB
				(\ref{eq:MCB}).
				Compute the p-value $p_{global}$ for this global calibration test.
				\Statex \vspace{0.35ex}\hrule \vspace{0.35ex}
				\Statex \textbf{Decision (global):}
				Reject the null hypothesis of no global shift if $p_{global}<\alpha_{global}$.
				A rejection indicates a global level shift.
				\Statex \vspace{0.35ex}\hrule \vspace{0.35ex}
				\State \textbf{Local level shift test (LMCB).}
				Apply a modified version of Algorithm~\ref{alg:autocalibration_test} to
				$\mathcal{T}_{new}$ that uses the local component LMCB (\ref{eq:LMCB}) in place
				of MCB.
				Compute the p-value $p_{local}$ for this local calibration test.
				\Statex \vspace{0.35ex}\hrule \vspace{0.35ex}
				\Statex \textbf{Decision (local):}
				Reject the null hypothesis of no local shift if $p_{local}<\alpha_{local}$.
				A rejection indicates local (cohort-level) shifts beyond any global level
				shift.
			\end{algorithmic}
		\end{algorithm}
	}

\medskip
\noindent
If no shift in ranking is detected (failure to reject in Algorithm~\ref{alg:gini_ranking_test}) and no local
level shift is found (failure to reject Part~(b) of Algorithm~\ref{alg:autocalibration_two_step}), yet
a global level shift is identified (rejection in Part~(a) of Algorithm~\ref{alg:autocalibration_two_step}),
the practitioner may decide to redeploy the existing model $\hat{\mu}_{old}(\cdot)$ with a balance-correction (positive affine transformation on the link scale).
This way maintaining the overall model structure and prior interpretability
while addressing the identified global shift, without requiring a granular
model update.
%
%
%
\subsection{Illustration example}\label{subsec:illustration_example}
To provide a simple illustrative example of the above-mentioned monitoring
framework, we use the de facto ``\textit{hello world}'' dataset for non-life
insurance pricing, the {\tt freMTPL2freq} dataset of
\cite{dutang_casdatasets_2018}.
\footnote{A cleaned version can be downloaded from
	\url{https://aitools4actuaries.com/}.
}
It is a well-known French motor third-party liability (MTPL) claim frequency
dataset that is widely used in the actuarial literature for benchmarking and
interpreting new methods.
Since the dataset is already well documented in the literature, we briefly
summarize where the data exploration, preprocessing, and model-fitting steps
can be found.
For data exploration, we refer to the tutorial by \cite{noll_case_2020}.
For data cleaning, feature engineering, and train/test split, we follow
\cite{wuthrich_statistical_2023} (Appendix B, Sec.~5.3.4, and Listing~5.2,
respectively).
A summary of the dataset characteristics is provided in Table
\ref{tab:dataset_characteristics}.

\noindent
We fit on the learning sample the same Poisson GLM with log-link as the \texttt{GLM3} model from
\cite{wuthrich_statistical_2023} (Sec.~5.3.4) using the scikit-learn API
\cite{sklearn_api} with the Newton-Cholesky solver.
This model uses all available categorical and numerical covariates
(\texttt{Area}, \texttt{VehGas}, \texttt{VehBrand}, \texttt{Region},
\texttt{VehPower}, \texttt{VehAge}, \texttt{DrivAge}, \texttt{BonusMalus},
\texttt{Density}).
The driver-age effect is modeled by normalized polynomial and logarithmic
terms, and interactions between driver age and the bonus-malus score are
included.

\noindent
For the illustration of our \textit{model monitoring} framework, we do not use the original
response variable \texttt{ClaimNb}.
Instead, we generate a new synthetic claim count dataset by drawing Poisson
responses with means equal to the \texttt{GLM3} predictions multiplied by the
exposures.
This way, the dataset preserves the original covariate and exposure
distributions, so it remains realistic while the response variable stays close
to the original one -- this also excludes a {\it virtual concept drift}.
This creates a controlled environment in which the true data-generating process
is known and the performance of our monitoring framework can be reliably
evaluated.
Henceforth, we denote \texttt{GLM3} as the \emph{true model} $\mu(\cdot)$.
The characteristics of the synthetic dataset are also summarized in Table
\ref{tab:dataset_characteristics} (lower part).
\begin{table}[h]
	\centering
	\caption{Dataset characteristics.}
	\label{tab:dataset_characteristics}
	\begin{tabular}{lcc}
		\toprule
		\textbf{Characteristic}             & \textbf{Learning set $\mathcal{L}$} & \textbf{Test set $\mathcal{T}$} \\
		\midrule
		Number of policies                  & 610,206                             & 67,801                          \\
		Total exposure (years)              & 322,392                             & 35,967                          \\
		\midrule
		\multicolumn{3}{l}{\textbf{Response summary (original cleaned data)}}                                       \\
		\midrule
		Number of claims                    & 23,738                              & 2,645                           \\
		Average frequency                   & 7.36\%                              & 7.35\%                          \\
		Minimal number of claims per policy & 0                                   & 0                               \\
		Maximal number of claims per policy & 5                                   & 5                               \\
		\midrule
		\multicolumn{3}{l}{\textbf{Response summary (synthetic data)}}                                              \\
		\midrule
		Number of claims                    & 23,687                              & 2,587                           \\
		Average frequency                   & 7.35\%                              & 7.19\%                          \\
		Minimal number of claims per policy & 0                                   & 0                               \\
		Maximal number of claims per policy & 4                                   & 4                               \\
		\bottomrule
	\end{tabular}
\end{table}

\noindent
On this synthetic dataset, we fit a new Poisson GLM with log-link.
By omitting the density of inhabitants, \texttt{Density}, and the interactions
between driver age and the bonus-malus score, this fitted model
$\hat{\mu}(\cdot)$ relies on a slightly different covariate set and structure
compared to the \textit{true model} $\mu(\cdot)$.
This new GLM model $\hat{\mu}(\cdot)$ effectively serves as a stand-in for a
real-world model that an insurer might use after fitting to historical data
without access to the true underlying model, that is, the insurer's model does
not access all the risk factors because they might not be available.
As a benchmark, we also fit a null model $\bar{\mu}$ (intercept only) on the
learning sample and consider the saturated model (perfect fit).
We summarize the Poisson deviance losses, Gini scores as well as average
predicted frequencies of all models on both learning and test sets in Table
\ref{tab:loss_gini_results}.
\begin{table}[h!]
	\centering
	\caption{Deviance losses in $10^{-2}$ and Gini scores on learning set $\mathcal{L}$ and test set $\mathcal{T}$.}
	\label{tab:loss_gini_results}
	\begin{tabular}{lccccccc}
		\toprule
		      &                        & \multicolumn{2}{c}{Poisson deviance loss} & \multicolumn{2}{c}{Gini score} & \multicolumn{2}{c}{Avg.\ freq.}                                                        \\
		\cmidrule(lr){3-4} \cmidrule(lr){5-6} \cmidrule(lr){7-8}
		Model &                        & $\mathcal{L}$                             & $\mathcal{T}$                  & $\mathcal{L}$                   & $\mathcal{T}$ & $\mathcal{L}$ & $\mathcal{T}$        \\
		\midrule
		(0)   & Saturated model        & 0.000                                     & 0.000                          & 1.000                           & 1.000         & 7.35\%        & \color{blue}{7.19\%} \\
		(1)   & Null model $\bar{\mu}$ & 45.174                                    & 44.117                         & 0.000                           & 0.000         & 7.35\%        & 7.35\%               \\
		(2)   & True model $\mu$       & 42.988                                    & 41.932                         & 0.280                           & 0.291         & 7.36\%        & 7.40\%               \\
		(3)   & GLM model $\hat{\mu}$  & 42.987                                    & 41.957                         & 0.281                           & 0.289         & 7.35\%        & \color{blue}{7.39\%} \\
		\bottomrule
	\end{tabular}
\end{table}

\begin{figure}[htb!]
	\centering
	\includegraphics[width=0.7\textwidth, trim={10pt 10pt 10pt 50pt}, clip]{
		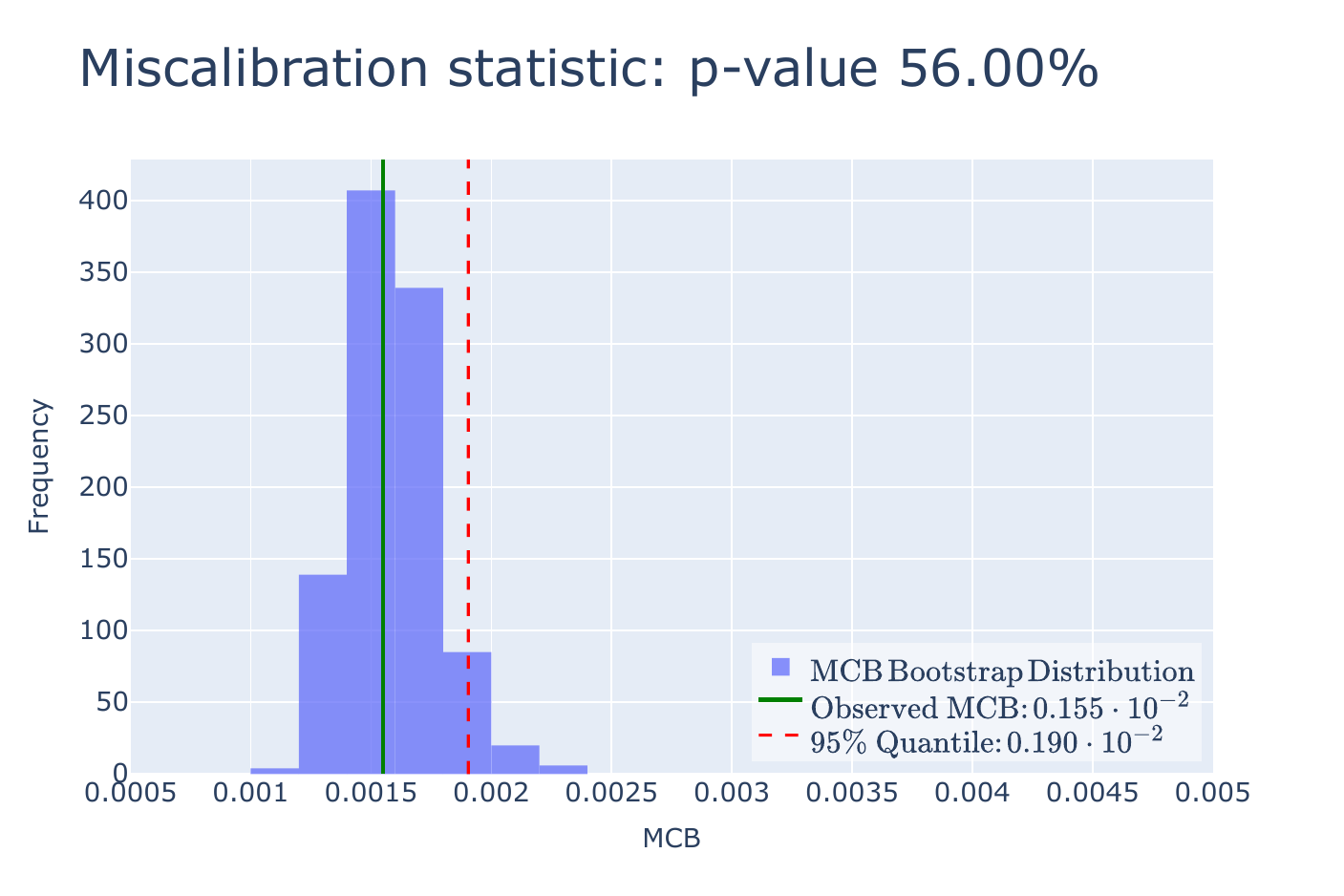}
	\caption{Auto-calibration test on the initial test set $\mathcal{T}$ based on the miscalibration statistic MCB: $p$-value $= 0.56$.}
	\label{fig:autocalibration_inital_test_illustration}
\end{figure}

\noindent
We observe that, as expected given the fairly large learning set $\mathcal{L}$, and a structure close to the true model,
the fitted GLM $\hat{\mu}(\cdot)$ closely approximates
the true model $\mu(\cdot)$ in terms of both deviance loss and Gini score on
the test set $\mathcal{T}$, indicating fairly good generalization performance.
Consistent with practical experience, the observed claim frequency on the
learning set (7.35\%) matches the predicted frequency well (a GLM with
canonical link satisfies the balance property), whereas on the test set the
observed claim frequency (7.19\%) differs slightly from the predicted
frequencies of the fitted model $\hat{\mu}(\cdot)$ (7.39\%).
Despite this imperfect prediction of the global frequency on the test set, the
auto-calibration test based on Algorithm~\ref{alg:autocalibration_test},
applied to the test set $\mathcal{T}$, does not indicate significant
miscalibration of the fitted model $\hat{\mu}(\cdot)$ ($p$-value $= 0.56$; see
Figure~\ref{fig:autocalibration_inital_test_illustration}).
This correctly indicates that the deviation at the global level is compatible
with statistical noise (irreducible risk), which is plausible because the true
model frequency on the test set $\mathcal{T}$ is in fact very close to that of
the fitted model $\mu(\cdot)$ (7.40\%).

\noindent
Furthermore, decomposing the overall miscalibration statistic MCB $= 0.155 \cdot 10^{-2}$
into its global (GMCB) and local (LMCB) components, see
equation~\eqref{eq:MCB_into_GMCB_LMCB}, shows that most of the (small)
miscalibration is driven by local effects (GMCB $= 0.005 \cdot 10^{-2}$, LMCB $= 0.150 \cdot 10^{-2}$),
rather than by a global level shift, further supporting the above conclusion.

\medskip
\noindent
In the following subsections, we illustrate the \textit{model monitoring} framework by
creating and analyzing scenarios that simulate \textit{concept drift} through rank shifts
as well as a global level shift.

\subsubsection{Concept Drift Scenarios Induced by Rank Shifts}
The \textit{true model} depends on several covariates, including the driver's
age, \texttt{DrivAge}, which we use to simulate \textit{concept drift} induced
by rank shifts.
To illustrate the monitoring framework, we construct datasets
$\mathcal{T}^{j}_\text{rank}$ for $j \in \{0,1,2,3\}$ by augmenting the
predictions of the \textit{true model} $\mu$ for different age groups and
generating new claim count observations.
In this way, we generate datasets that exhibit \textit{concept drift} scenarios
of increasing magnitude $j$ while preserving knowledge of the underlying new
\textit{true model} $\mu^{j}_\text{rank}$.
The \textit{true models} in the scenarios are defined as follows:
\begin{equation*}
	\mu^{j}_\text{rank} =
	\begin{cases}
		\mu     &
		\text{if } \texttt{DrivAge} \le 30, \\[4pt]
		\mu  \left(
		1 + \dfrac{\texttt{DrivAge} - 30}{\overline{\texttt{DrivAge}}} \, s_j
		\right) &
		\text{if } \texttt{DrivAge} > 30,
	\end{cases}
	\quad\text{with}\quad
	s_j =
	\begin{cases}
		0   & \text{for } j = 0, \\
		0.3 & \text{for } j = 1, \\
		0.5 & \text{for } j = 2, \\
		0.8 & \text{for } j = 3,
	\end{cases}
\end{equation*}
\noindent
This means that for drivers older than 30 years, we adjust the true claim
frequency $\mu$ by a scaling factor that increases linearly with the driver's
age.
Consequently, the older the driver, the larger the deviation from the original
true model $\mu$.
By simulating claims from the new \textit{true model} $\mu^{j}_\text{rank}$,
this mimics a situation where changes in driving behavior within demographic
groups alter their claim frequencies over time, even though the portfolio
composition itself does not change.
The scaling factors for the four scenarios are such that in scenario $j=0$ the
new data generating process is identical to the original one, while in
scenarios $j=1,2,3$ we induce progressively more severe \textit{concept drift}.

\begin{figure}[htb!]
	\begin{subfigure}[t]{0.49\textwidth}
		\centering
		\includegraphics[width=\textwidth, trim={10pt 8pt 20pt 55pt}, clip]{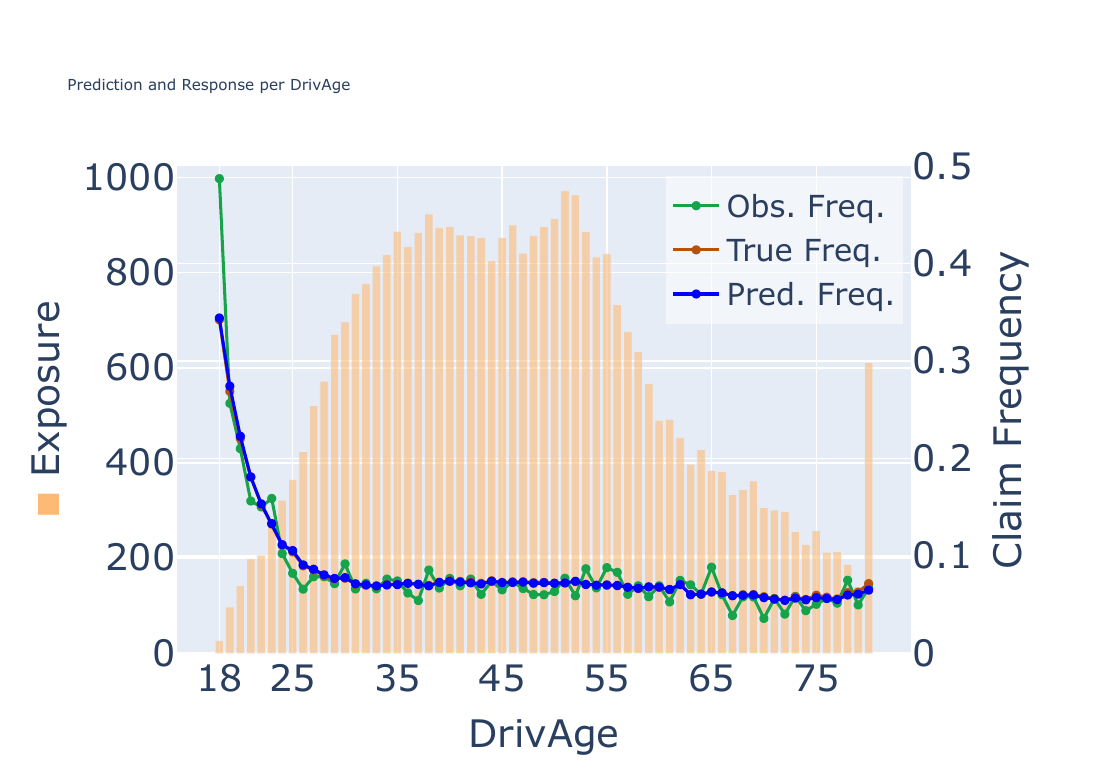}
		\captionsetup{justification=centering}\caption{Scenario 0 \
			$\begin{aligned}
					\text{Gini true model } \mu        = \phantom{-}   & 0.2910 \\[-4pt]
					\text{Gini GLM model } \hat{\mu} = \phantom{-}     & 0.2893 \\[-4pt]
					p\text{-value}                       = \phantom{-} & 0.6240 \\[-4pt]
					z\text{-statistic}                  = \phantom{-}  & 0.4902
				\end{aligned}$}
	\end{subfigure}
	\begin{subfigure}[t]{0.49\textwidth}
		\centering
		\includegraphics[width=\textwidth, trim={10pt 8pt 20pt 55pt}, clip]{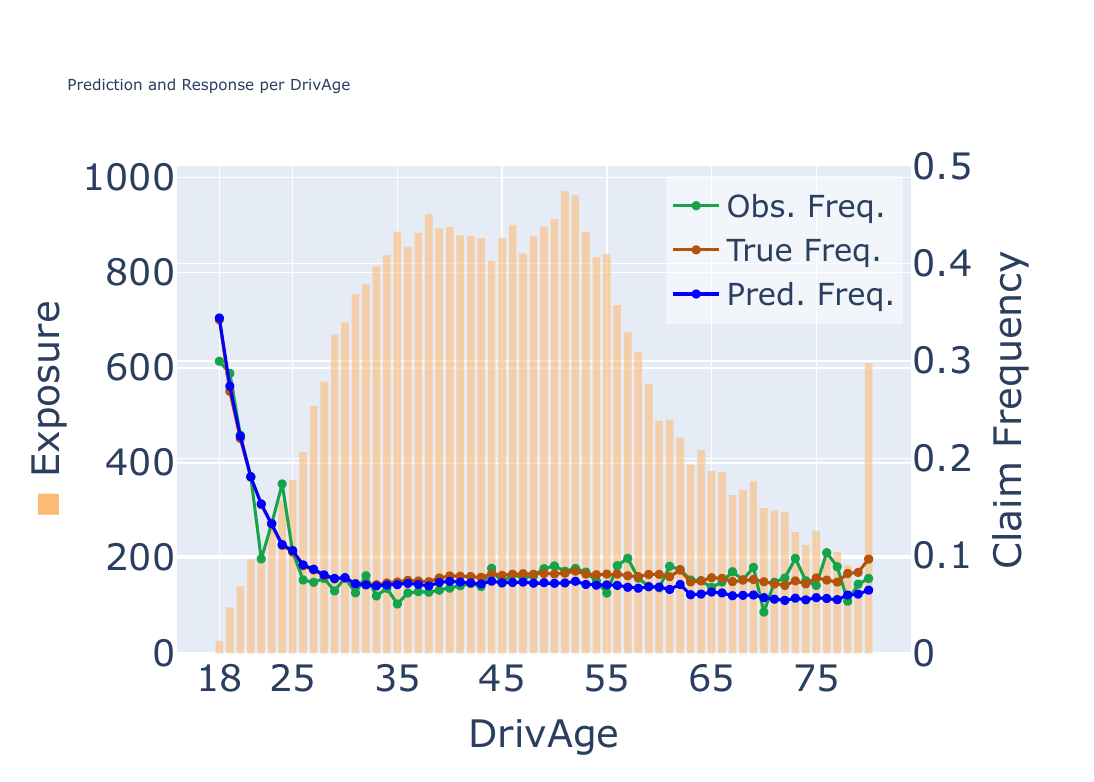}
		\captionsetup{justification=centering}\caption{Scenario 1 \
			$\begin{aligned}
					\text{Gini true model } \mu        = \phantom{-}   & 0.2750 \\[-4pt]
					\text{Gini GLM model } \hat{\mu} = \phantom{-}     & 0.2683 \\[-4pt]
					p\text{-value}                       = \phantom{-} & 0.1979 \\[-4pt]
					z\text{-statistic}                  = -            & 1.2875
				\end{aligned}$}
	\end{subfigure}

	\begin{subfigure}[t]{0.49\textwidth}
		\centering
		\includegraphics[width=\textwidth, trim={10pt 8pt 20pt 55pt}, clip]{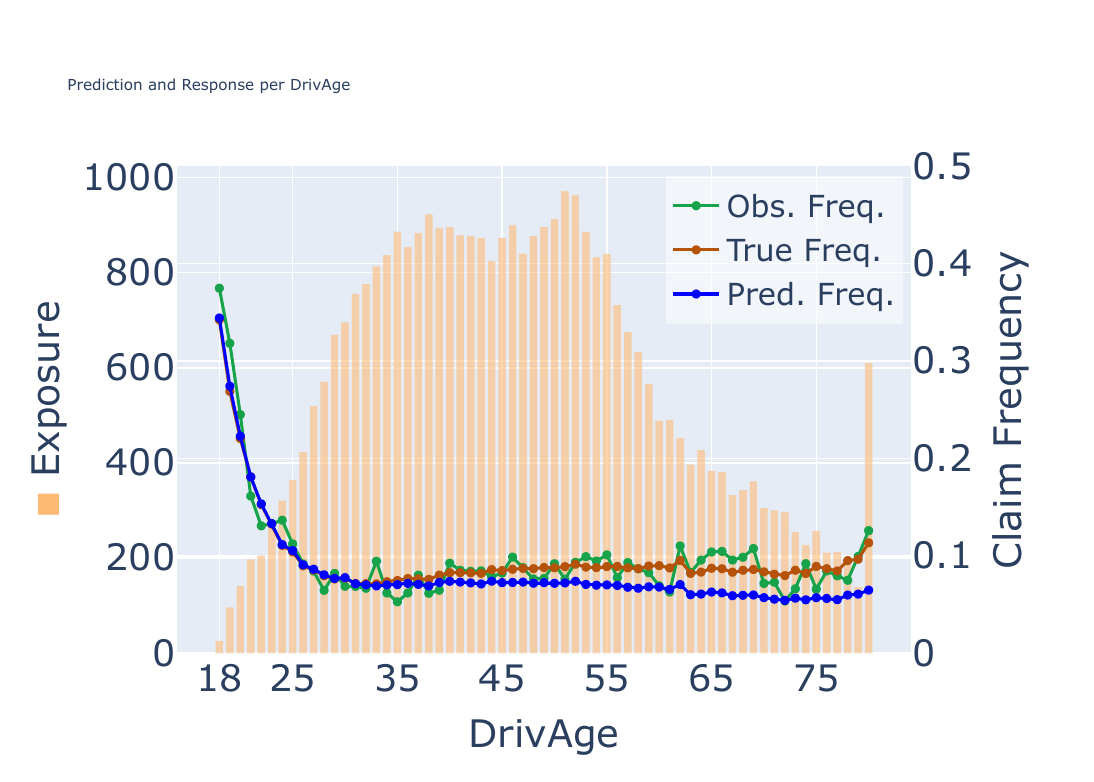}
		\captionsetup{justification=centering}\caption{Scenario 2 \
			$\begin{aligned}
					\text{Gini true model } \mu        = \phantom{-}   & 0.2720 \\[-4pt]
					\text{Gini GLM model } \hat{\mu} = \phantom{-}     & 0.2562 \\[-4pt]
					p\text{-value}                       = \phantom{-} & 0.0210 \\[-4pt]
					z\text{-statistic}                  = -            & 2.3087
				\end{aligned}$}
	\end{subfigure}
	\begin{subfigure}[t]{0.49\textwidth}
		\centering
		\includegraphics[width=\textwidth, trim={10pt 8pt 20pt 55pt}, clip]{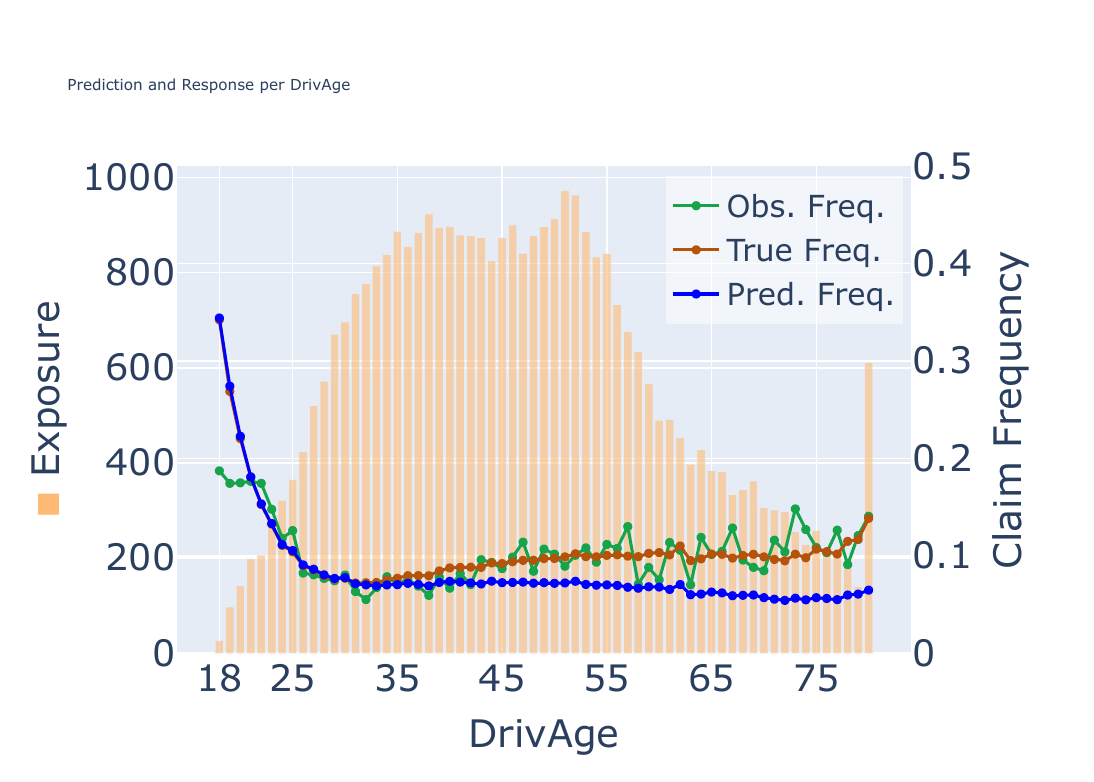}
		\captionsetup{justification=centering}\caption{Scenario 3 \
			$\begin{aligned}
					\text{Gini true model } \mu        = \phantom{-}   & 0.2860 \\[-4pt]
					\text{Gini GLM model } \hat{\mu} = \phantom{-}     & 0.2549 \\[-4pt]
					p\text{-value}                       = \phantom{-} & 0.0157 \\[-4pt]
					z\text{-statistic}                  = -            & 2.4169
				\end{aligned}$}
	\end{subfigure}
	\caption{Illustrative example showing the effect of induced concept drift by changing the driver age effect.}
	\label{fig:illustration_example}
\end{figure}

\noindent
Figure \ref{fig:illustration_example} visualizes the changes introduced by this
procedure.
In each plot, bars represent exposure per age group, the green line shows
observed claim frequency, the true marginal claim frequencies
$\mu^{j}_\text{rank}$ are represented as a red line and the predicted claim
frequencies from the historical GLM $\hat{\mu}$ are shown as a blue line.
The model's predicted claim frequency $\hat{\mu}$ remains unchanged, because
policyholder features do not change, this way mimicking a scenario in which we
do not observe covariate drift but a pure \textit{concept drift}.
We note that for increasing $j$, the scenarios lead to an increasing U-shape in
the marginal frequency as a function of the driver's age variable, and as a
result, this leads to an increasingly wrong risk ranking between younger (below
30) and older drivers.

\noindent
By applying the Gini based ranking drift test from Algorithm~\ref{alg:gini_ranking_test},
we obtain $p$-values and $z$-statistics that indicate whether the risk ranking
performance of the historical model $\hat{\mu}$ has deteriorated on the new
datasets.
Moreover, since we work in a controlled simulation setting, we can also
compute, for each scenario $j$, the Gini score implied by the generated
observations and the corresponding new true model $\mu^{j}_\text{rank}$.
This allows us to directly quantify the realized ranking performance under
\textit{concept drift} and compare it to the performance of the historical
model.

\noindent
The results reveal that as expected, in Scenario 0 (no \textit{concept drift}), the
$p$-value is high ($0.6240$) suggesting no significant change in ranking
performance.
In contrast, for the more severe \textit{concept drift} introduced in Scenarios
1 to 3, we observe decreasing $p$-values of $0.1979$, $0.0210$, and $0.0157$,
respectively, indicating increasing statistical evidence of model
deterioration.
Furthermore, as expected, the corresponding $z$-statistics become more
negative, moving from $-1.2875$ in Scenario 1 to $-2.4169$ in Scenario 3,
indicating larger drops in ranking performance.

\noindent
It is important to note that in this monitoring context, the trade-off between Type I
and Type II errors is asymmetric.
A Type I error (false alarm) triggers an unnecessary model review or update,
which requires operational effort but preserves model performance.
In contrast, a Type II error (missed detection) allows a degraded model to
remain in production, potentially leading to financial loss or wrong decisions.
Therefore, practitioners may prefer to set a higher significance level $\alpha$
to minimize the risk of missing a necessary update.

\begin{figure}[htb!]
	\begin{subfigure}[t]{0.49\textwidth}
		\centering
		\includegraphics[width=\textwidth, trim={10pt 8pt 20pt 55pt}, clip]{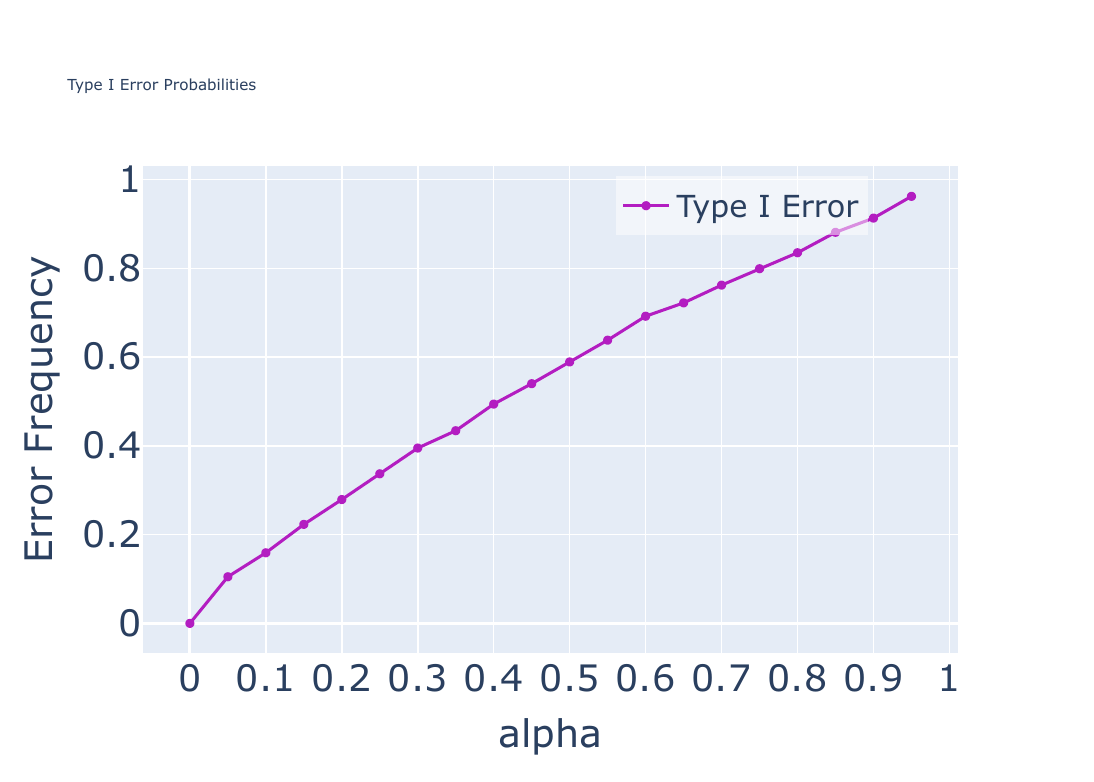}
		\captionsetup{justification=centering}\caption{Type I Error Probabilities}
	\end{subfigure}
	\begin{subfigure}[t]{0.49\textwidth}
		\centering
		\includegraphics[width=\textwidth, trim={10pt 8pt 20pt 55pt}, clip]{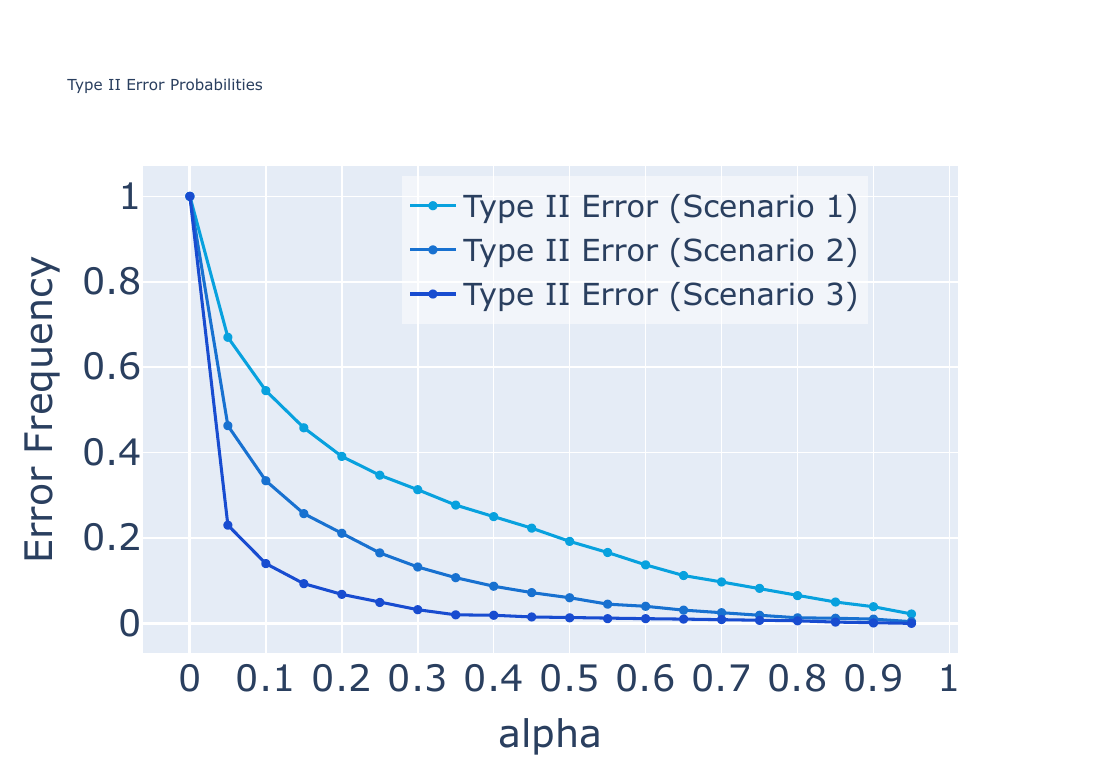}
		\captionsetup{justification=centering}\caption{Type II Error Probabilities}
	\end{subfigure}
	\caption{Type I and Type II errors per significance level $\alpha$ and scenario $j \in \{0,1,2,3\}$.}
	\label{fig:type1_type2_error_probabilities}
\end{figure}

\noindent
We estimate the Type I error rate of the monitoring test by simulating 1,000 datasets $\mathcal{T}^{0}_\text{rank}$ under Scenario 0 (no \textit{concept drift}) and
calculating the proportion of times the ranking drift test incorrectly signals drift at various significance levels $\alpha$ (see Figure
\ref{fig:type1_type2_error_probabilities}(a)).
Similarly, to estimate the Type II error rate, we simulate 1,000 datasets
$\mathcal{T}^{j}_\text{rank}$ for each scenario $j \in \{1,2,3\}$ where concept
drift is present.
We then calculate the proportion of times the ranking drift test fails to
detect drift across different significance levels $\alpha$, as shown in Figure
\ref{fig:type1_type2_error_probabilities}(b).

\noindent
As expected, the Type I error rate increases with the significance level
$\alpha$.
In contrast, the Type II error rate declines as $\alpha$ increases and as the
magnitude of \textit{concept drift} grows (Scenario 1 through Scenario 3).
The results further show that $\alpha = 0.05$ yields an undesirably high Type
II error rate in this monitoring context.
Depending on the magnitude of \textit{concept drift} the insurer aims to
detect, a higher significance level such as $\alpha = 0.32$ (capturing
degradation exceeding one standard deviation) or even higher seems more
appropriate for this monitoring framework.
\subsubsection{Concept Drift Scenarios Induced by Global Level Shifts}
\begin{figure}[H]
	\begin{subfigure}[t]{0.49\textwidth}
		\centering
		\includegraphics[width=\textwidth, trim={10pt 8pt 20pt 55pt}, clip]{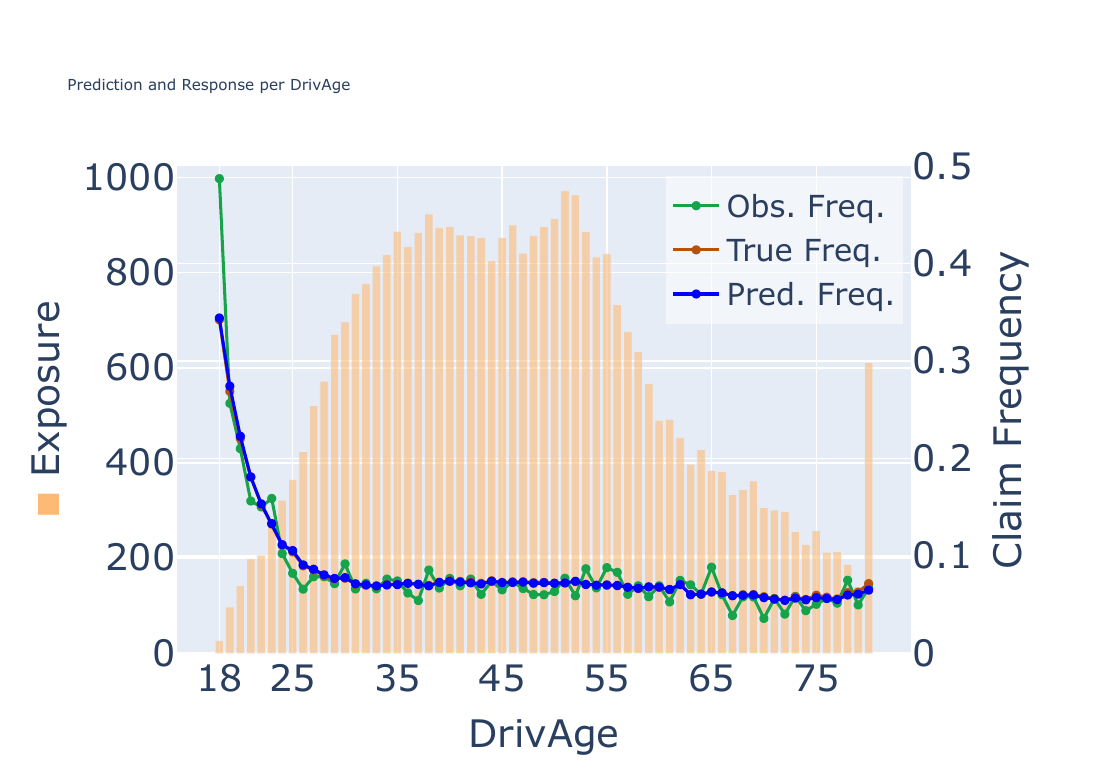}
		\captionsetup{justification=centering}\caption{Scenario 0 \
			$\begin{aligned}
					\text{Gini true model } \mu        = \phantom{-}        & 0.2910 \\[-4pt]
					\text{Gini GLM model } \hat{\mu} = \phantom{-}          & 0.2893 \\[-4pt]
					p\text{-value Gini}                       = \phantom{-} & 0.6240 \\[-4pt]
					\\
					\text{MCB } \hat{\mu} \text{ in } 10^{-2} = \phantom{-} & 0.1550 \\[-4pt]
					p\text{-value MCB}                       = \phantom{-}  & 0.5600 \\[-4pt]
					p\text{-value GMCB}                       = \phantom{-} & 0.4040 \\[-4pt]
					p\text{-value LMCB}                       = \phantom{-} & 0.6360 \\[-4pt]
				\end{aligned}$}
	\end{subfigure}
	\begin{subfigure}[t]{0.49\textwidth}
		\centering
		\includegraphics[width=\textwidth, trim={10pt 8pt 20pt 55pt}, clip]{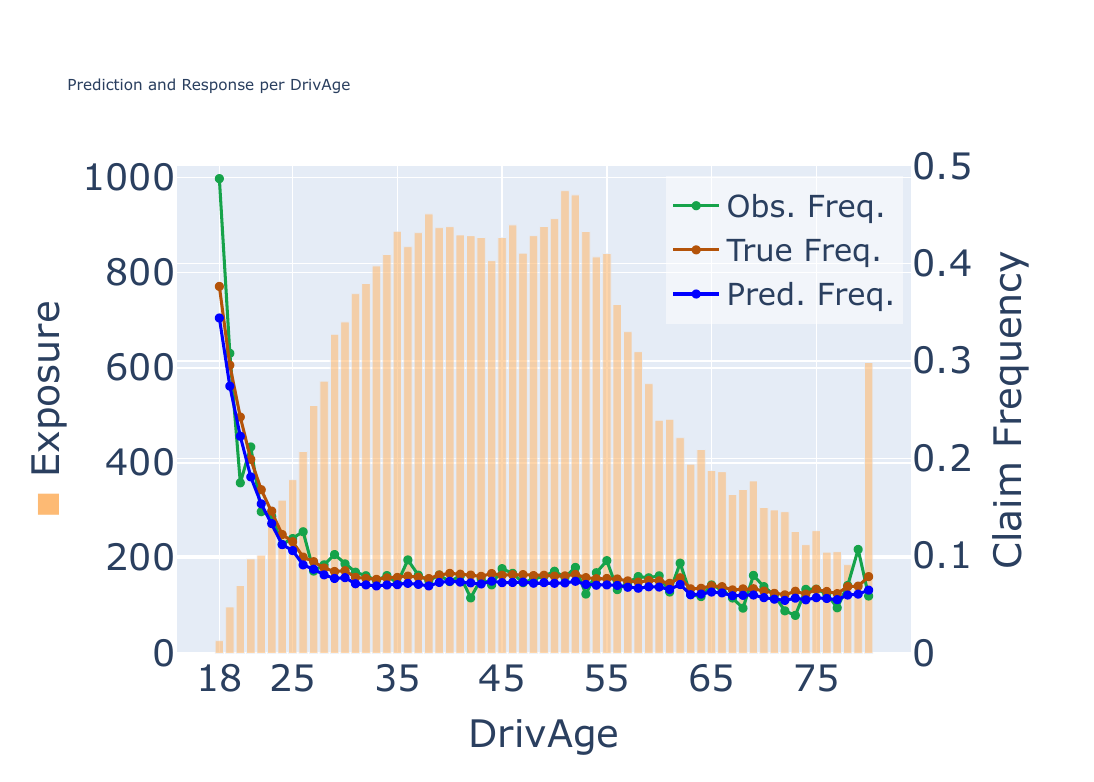}
		\captionsetup{justification=centering}\caption{Scenario 1 \
			$\begin{aligned}
					\text{Gini true model } \mu        = \phantom{-}        & 0.2887 \\[-4pt]
					\text{Gini GLM model } \hat{\mu} = \phantom{-}          & 0.2879 \\[-4pt]
					p\text{-value Gini}                       = \phantom{-} & 0.7159 \\[-4pt]
					\\
					\text{MCB } \hat{\mu} \text{ in } 10^{-2} = \phantom{-} & 0.2276 \\[-4pt]
					p\text{-value MCB}                       = \phantom{-}  & 0.0040 \\[-4pt]
					p\text{-value GMCB}                       = \phantom{-} & 0.0000 \\[-4pt]
					p\text{-value LMCB}                       = \phantom{-} & 0.2890 \\[-4pt]
				\end{aligned}$}
	\end{subfigure}
	\caption{Illustrative example showing the effect of induced global level drift.}
	\label{fig:global_level_shift_illustration}
\end{figure}

To illustrate \textit{concept drift} induced by global level shifts, we construct a new dataset,
denoted as $\mathcal{T}_\text{global}$, by applying a constant scaling factor to the predictions of the \textit{true model} $\mu$ across all policies:
\begin{equation*}
	\mu_\text{global} =
	\mu \, (1 + s_\text{global}),
	\quad\text{with}\quad
	s_\text{global} = 0.1.
\end{equation*}
This adjustment raises the true claim frequency on the test set $\mathcal{T}$
from $7.4\%$ to $8.1\%$ on $\mathcal{T}_\text{global}$, thereby simulating a
trend-driven increase that is independent of specific covariate values.
We visualize this global shift in Figure
\ref{fig:global_level_shift_illustration} by comparing the true model (red
line) against the historical fitted model (blue line) with respect to the
driver age feature.

\noindent
As anticipated, since the rank ordering of the true and fitted models remains invariant under a global scalar shift,
the Gini based ranking drift test (Algorithm~\ref{alg:gini_ranking_test}) detects no significant deterioration in performance ($p\text{-value} = 0.7159$).
Conversely, the auto-calibration test
(Algorithm~\ref{alg:autocalibration_test}) correctly flags a significant
miscalibration of the fitted model $\hat{\mu}(\cdot)$ on the new dataset
$\mathcal{T}_\text{global}$ ($p\text{-value} = 0.0040$; see
Figure~\ref{fig:global_level_shift_illustration}(b)).
Furthermore, the decomposition of the miscalibration statistic (MCB $= 0.2276
	\cdot 10^{-2}$) confirms that the drift is driven by the global component (GMCB
$p\text{-value} < 0.001$), whereas the local component remains statistically
insignificant (LMCB $p\text{-value} = 0.2890$).
\noindent
%
%
%
\subsection{Practical Considerations and Pitfalls}\label{subsec:practical_considerations}
In this section, we discuss practical considerations for implementing the
proposed framework and summarize common pitfalls.
We first list actionable setup recommendations, then a short checklist of basic
pitfalls, and finally detail advanced pitfalls and their mitigations.

\bigskip

\noindent
\textbf{Practical considerations (setup and process):}

\medskip
\noindent
\underline{\textit{Significance level and monitoring frequency:}}
Regarding the choice of the significance level $\alpha$, for the Gini ranking
test (Algorithm~\ref{alg:gini_ranking_test}) as well as for the global
($\alpha_{global}$) and local ($\alpha_{local}$) level shift tests
(Algorithm~\ref{alg:autocalibration_two_step}), we deliberately choose not to
recommend adopting a fixed level such as the commonly used $\alpha=5\%$.
The reason is that, in a \textit{model monitoring} context, the decision to
replace a model typically involves a trade-off between performance
considerations vs.~stability and implementation costs, which is highly
dependent on the specific business context.
As mentioned above, the annual monitoring frequency is only an example; in
reality, the monitoring cycle may vary depending on the model's purpose as well
as the business, implementation, and regulatory contexts.
So the choice of the significance level $\alpha$ should reflect the Type I
vs.~Type II error trade-offs that are specific to the given context.
\\
\\
\noindent
\underline{\textit{Regarding the types of real concept drift:}}
Depending on the detection of \textit{real concept drift type} (see Section
\ref{sec:virtual_concept_drift}), one may use different holdout samples to
estimate the mean and standard deviation of the Gini score based on the
training period.
For example, in the case of \textit{sudden drift}, one might use a holdout
sample consisting only of the most recent training year to estimate the mean
and standard deviation of the Gini score.
In cases of \textit{gradual drift} or \textit{incremental drift}, one might
compute separate mean and standard deviation estimates for each training year's
holdout set, and conducting the hypothesis test separately for each year using
its corresponding estimates.
If \textit{recurrent drift} due to seasonality is already known (e.g.,
weather-related monthly patterns), one should apply the above approach to
datasets restricted to the relevant seasonal periods.

\bigskip
\noindent
Some pitfalls are often overlooked: while they matter less in model comparison
settings, they can have a material impact in model-monitoring applications.

\medskip
\noindent
\textbf{Pitfalls:}
\begin{itemize}
	\item \underline{\textit{Holdout $\mathcal{T}$}}:
	      Using the training data $\mathcal{L}$ from the model development period instead of a separate holdout sample $\mathcal{T}$ generally leads to under-estimated variability and, consequently, inflated Type~I error rates.
	\item \underline{\textit{New test data $\mathcal{T}_{new}$}}:
	      For the new period data $\mathcal{T}_{new}$, it is important to use a sample that is comparable in size and covariate distribution to the holdout set $\mathcal{T}$ on which the bootstrap estimates of the Gini mean and standard deviation are based on.
	      While detecting covariate drift (e.g., changes in portfolio composition over
	      time) is also important for insurers, such analyses lie beyond the scope of
	      this work.
	\item \underline{\textit{Metric implementation}}:
	      Under-estimating the impact of different implementations of the Gini score.
	      Particularly in the presence of prediction ties and case weights this can lead
	      to misleading conclusions.
	      There exist multiple implementations of the Gini score, and these differences
	      can have a substantial effect on the resulting performance measures.
	      We therefore recommend using a consistent implementation throughout the
	      monitoring process.
	      Further details on the approach advocated in this manuscript are provided in
	      \cite{brauer_wuethrich_gini_2025}.
	\item \underline{\textit{Implications of time splitting.}}
	      The following aspect of data preparation pipelines (ETL pipelines) for
	      \textit{model monitoring} is often under-estimated.
	      In claim frequency modeling, one typically works with datasets $\mathcal{D}$ in
	      which each row represents a specific time period for a policyholder.
	      These datasets are frequently transformed by splitting single rows into
	      multiple rows, each corresponding to a shorter time period for the same
	      policyholder, yielding a time-split version $\mathcal{D}'$ (with unchanged
	      covariates, adjusted exposure, and indicators for whether a claim occurred in
	      each sub-interval).
	      One motivation for such time splitting is to simplify reporting: having at most
	      one claim per row allows the claim date and other response information to be
	      stored directly, which is difficult in traditional data structures when
	      multiple claims within a period are aggregated.
	      Another reason is that one considers annual data, and for contracts that are
	      renewed during the calendar year, one enters two different rows for the two
	      contract periods.
	      Such transformations preserve the average claim frequency, the total number of
	      claims, and the total exposure $\sum_{i=1}^{n} v_i$.
	      In particular, for a Poisson GLM, inspection of the score equations shows that,
	      because the sufficient statistics remain unchanged, the estimated coefficients
	      are identical whether the model is fitted to $\mathcal{D}$ or to
	      $\mathcal{D}'$.
	      \\
	      However, time splitting can substantially affect \textit{model monitoring} diagnostics.
	      Using the dataset and model from Section~\ref{subsec:illustration_example}, we
	      applied a time-period split such that each claim is represented by exactly one
	      row with an exposure of one day ($1/365$ of a year).
	      For rows with multiple claims, we created multiple one-day rows, each
	      containing exactly one claim, and assigned the remaining exposure to an
	      additional row with zero claims.
	      The fitted GLM coefficients remained unchanged up to numerical precision, but
	      the Gini score dropped from $0.2893$ to $0.2774$, and the Poisson deviance loss
	      increased from $41.957\cdot 10^{-2}$ to $120.580\cdot 10^{-2}$.
	      The dramatic change in the deviance loss is driven by the strong effect of the
	      weights on the log-likelihood, and the change in the Gini score is illustrated
	      by the CAP curves in Figure~\ref{fig:CAP_curve_time_split}.

	      \begin{figure}[htb!]
		      \begin{subfigure}[t]{0.48\textwidth}
			      \centering
			      \includegraphics[width=\textwidth, trim={10pt 10pt 10pt 55pt}, clip]{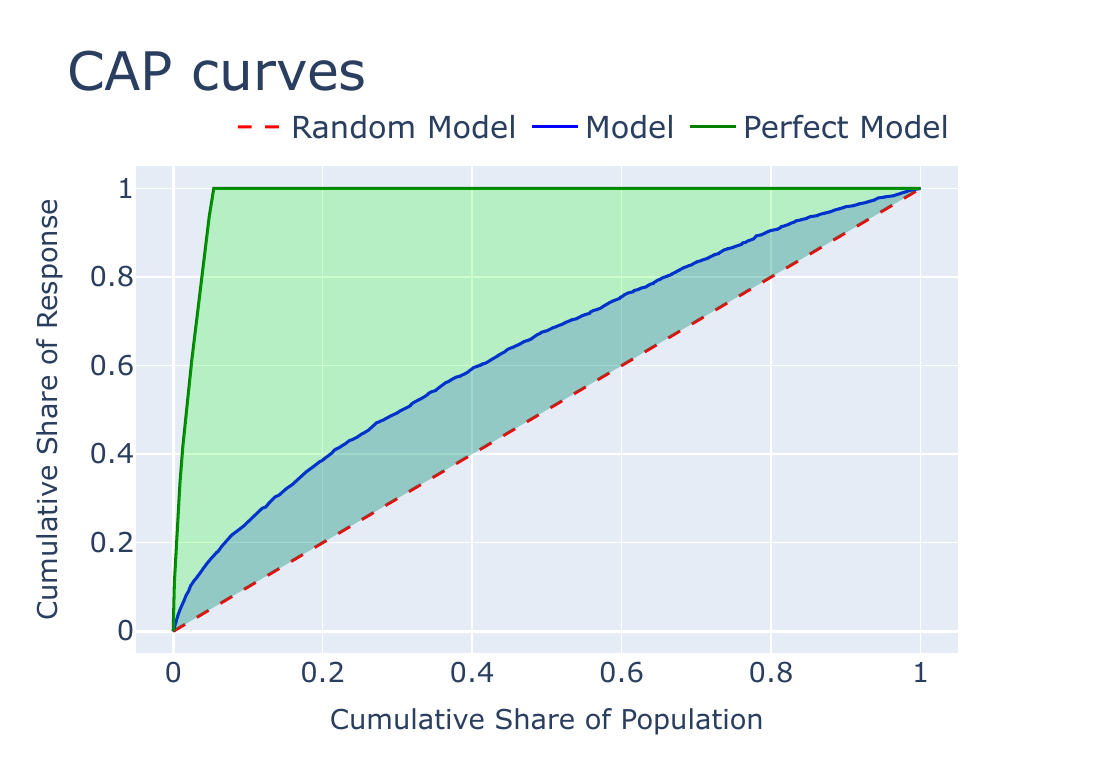}
			      \captionsetup{justification=centering}\caption{Before time-splitting. \
				      $\begin{aligned}
						      \text{Gini score } \hat{\mu} =                                & \; 0.2893 \\[-4pt]
						      \text{Poisson deviance loss }   \hat{\mu}                   = & \; 41.957 \\[-4pt]
					      \end{aligned}$}
		      \end{subfigure}
		      \begin{subfigure}[t]{0.48\textwidth}
			      \centering
			      \includegraphics[width=\textwidth, trim={10pt 10pt 10pt 55pt}, clip]{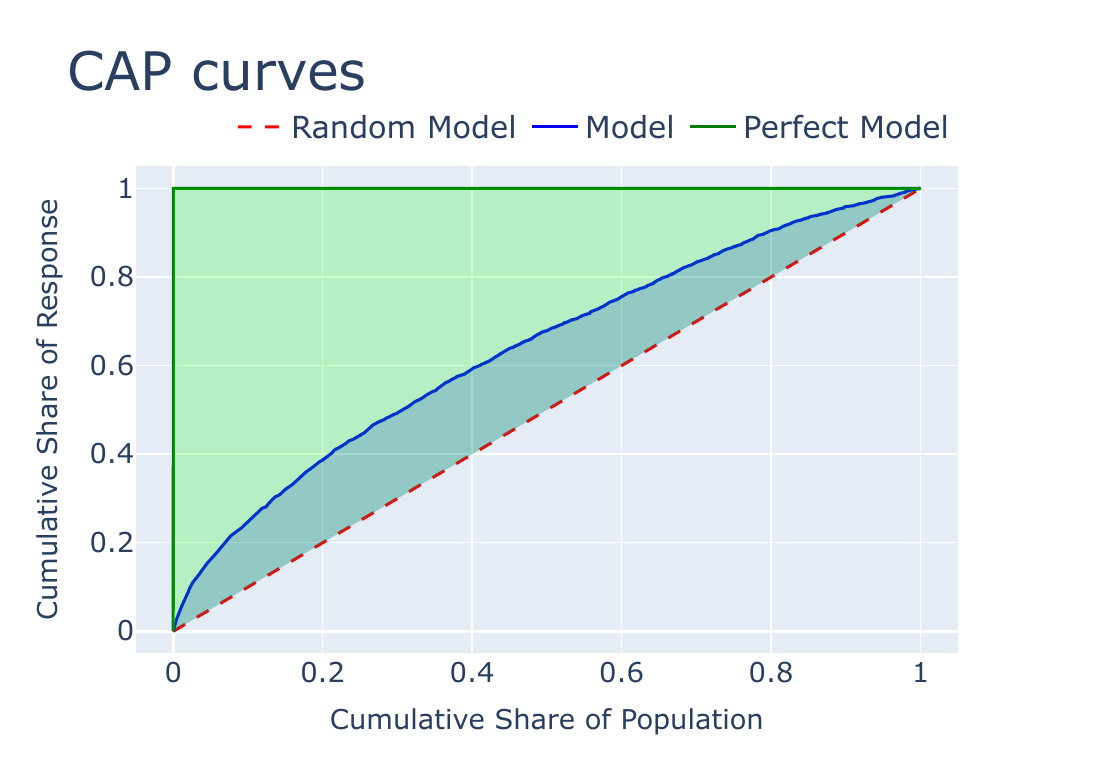}
			      \captionsetup{justification=centering}\caption{After time-splitting. \
				      $\begin{aligned}
						      \text{Gini score } \hat{\mu} =                                & \; \phantom{0} 0.2774 \\[-4pt]
						      \text{Poisson deviance loss }   \hat{\mu}                   = & \; 120.580            \\[-4pt]
					      \end{aligned}$}
		      \end{subfigure}
		      \caption{CAP curves before time-splitting (left) vs. after time-splitting (right).}
		      \label{fig:CAP_curve_time_split}
	      \end{figure}
	      \noindent
	      From Figure~\ref{fig:CAP_curve_time_split}, we observe that the model CAP curve
	      (blue) remains essentially unchanged.
	      This is because the time-splitting procedure only introduces additional ties in
	      the predictions, which are handled by the sorting and case-weight aggregation
	      scheme described in Definition~\ref{def:empirical_gini_score}, leaving the
	      average area under the model CAP curve invariant.
	      The change in the Gini score is instead driven by the alteration of the
	      \emph{best} CAP curve (green): time-splitting reduces the weights attached to
	      individual claim observations, which makes the best-model curve steeper and
	      increases the area in the denominator of the Gini score.
	      \\
	      These effects can materially distort the \textit{model monitoring} process, as both the
	      Gini score and the auto-calibration assessment via the deviance loss may then
	      lead to misleading conclusions.
	      Since time-period splitting is routinely applied for various purposes in large
	      ETL pipelines, this example highlights an important pitfall: seemingly minor
	      ETL changes that leave the model fit unchanged can nonetheless have a
	      substantial impact on downstream \textit{model monitoring} diagnostics.
	      \\
	      \\
	      \noindent
	      \textit{Recommendation.}
	      To avoid this pitfall, we recommend pre-aggregating the data before using it in
	      a \textit{model monitoring} context, at least at the policyholder level.
	      This pre-aggregation should be applied not only to the new-period data
	      $\mathcal{T}_{new}$, but also prior to creating the holdout set $\mathcal{T}$
	      from the model development period.
	      While this introduces a small additional computational step in the data
	      preparation pipeline, it simultaneously reduces model inference time because
	      the resulting datasets are smaller.

	      \bigskip

	      Remark that time splitting can also be problematic in a classical model
	      development set-up because if one partitions the available data at random into
	      training and validation data, there can be a leakage of information from one to
	      the other sample by the fact that the same policyholder may appear in both
	      samples, due to a time-splitting, e.g., caused by a contract renewal.

\end{itemize}

%% file: 01_Sec4_Conclusion_and_Outlook.tex
\section{Conclusion}\label{sec:conclusion}
\noindent
This paper provides a systematic examination of \textit{concept drift}
in non-life insurance pricing and a statistically grounded monitoring framework.
A comprehensive overview of the relevant literature on \textit{concept drift}
is provided and contextualized in the actuarial setting.
We derive the asymptotic distribution of the Gini score to enable valid
inference and hypothesis testing.
Building on this, we propose a standardized monitoring procedure that signals
when refitting is warranted due to degradation in ranking ability or
calibration, and illustrate its practical use on a modified real-world
portfolio in which we inject controlled levels of \textit{concept drift}.
We highlight implementation considerations and several pitfalls for model
monitoring and model comparison.
\\
\noindent
The described framework is model-agnostic and applies not just for GLMs but equally
to modern machine-learning models such as tree ensembles and neural networks.
In practice, the approach supports transparent and repeatable monitoring and
governance, helping prioritize refitting efforts where they create the most
value.
\\
\noindent
Methodologically, several extensions are promising and warrant exploration in
future research.
Different windowing designs and adaptive schemes could be investigated to
improve responsiveness and robustness.
Recurrent drift deserves special attention, particularly in long-term business.
In addition, combining multiple \textit{concept drift} detection methods with
dimensionality-reduction diagnostics could improve attribution and reveal the
drivers of drift.
While the focus of our work is on drift detection, future work could benchmark
drift-adaptation strategies for pricing, including windowing-based updates,
ensemble methods, and continual learning to maintain performance while
preserving valuable prior knowledge.

%% file: 01_Sec5_Appendices.tex
\begin{appendices}
	\section{Proof of Theorem \ref{thm:gini_asymptotic_normality}}\label{app:proof_gini_asymptotic_normality}
	\begin{proof}[Proof of Theorem \ref{thm:gini_asymptotic_normality} (Asymptotic Normality of the machine-learning Gini score)]
		\normalfont
		First, observe that for continuous marginal distributions of $Y$ and $\hat{\mu}$, the machine-learning Gini score $\widehat{G}_n(Y,\hat{\mu})$ in Definition \ref{def:empirical_gini_score} simplifies to $\widehat{G}_n(Y,\hat{\mu}) = \nicefrac{A}{B}$ because $A^\downarrow = A^\uparrow$.

		\medskip
		\noindent
		Moreover, both areas $A$ and $B$ can be represented as scaled empirical Gini indices based on the \textit{ordered Lorenz curve} described by \cite{frees_summarizing_2011} (these Gini indices differ from the machine-learning Gini scores).
		To adopt the notation of \cite{frees_summarizing_2011}, set the premiums to
		$\Pi(\mathbf{x}) \equiv 1$.
		Area $A$ corresponds to the Gini index computed from scores $S(\mathbf{x}) =
			\hat{\mu}(\mathbf{x})$, and area $B$ corresponds to the Gini index computed
		from scores $S(\mathbf{x}) = y_i$ (which asymptotically relates to an {\it
				ordered Lorenz curve} based on the true model $\mu(\mathbf{x})$).
		In this notation, the Gini indices equal $2A$ and $2B$, respectively.

		\medskip
		\noindent
		Consequently, the machine-learning Gini score can be expressed as the function $g(\cdot)$ (specifically, a quotient) of these two Gini indices, $g(2A, 2B)$.
		Applying the multivariate normality result for Gini indices (Theorem 5 in
		\cite{frees_summarizing_2011}) together with the multivariate Delta method
		(see, e.g., Equation (1.9) in \cite{wasserman_all_2006}) yields the asymptotic
		normality of the machine-learning Gini score.

		\medskip
		\noindent
		Finally, the general case with case weights $v_i$ follows by setting premiums $\Pi(\mathbf{x}) = v_i$ and scores $S(\mathbf{x}) = v_i\hat{\mu}(\mathbf{x})$ when using weighted losses.
	\end{proof}

	\section{Explicit Form for the Deviance Loss}\label{app:explicit_form_deviance_loss_gamma_poisson}
	To illustrate the deviance loss for practical applications, we provide its
	explicit form for the gamma and the Poisson EDF cases.
	\\
	\begin{example}[Deviance loss for gamma EDF]\label{ex:deviance_loss_gamma}\normalfont
		In the case of the gamma EDF case, given a dataset $\mathcal{D} = {\{\left(y_{i},
			\hat{\mu}_{i}, v_{i}\right)\}}_{i=1}^{n}$, the gamma deviance loss is given by
		\begin{equation*}
			S(\mathbf{y}, \hat{\boldsymbol{\mu}}, \mathbf{v}) = \frac{2}{\sum_{i=1}^{n} v_i} \sum_{i=1}^{n}
			\frac{v_i}{\varphi} \left(\frac{y_i - \hat{\mu}_i}{\hat{\mu}_i} +
			\log\left(\frac{\hat{\mu}_i}{y_i}\right)\right),
		\end{equation*}
		where $\varphi=1/\gamma>0$ is the dispersion parameter for gamma shape parameter $\gamma>0$.
		Usually, to make gamma deviance losses comparable across models, one sets
		$\varphi=1$.
		In a claim severity setting, $y_i$ denotes the observed average severity and
		the exposure $v_i \in \mathbb{N}$ denotes the claim count for policy $i$.
	\end{example}

	\medskip

	\begin{example}[Deviance loss for Poisson EDF]\label{ex:deviance_loss_poisson}\normalfont
		The Poisson deviance loss is given by
		\begin{equation*}
			S(\mathbf{y}, \hat{\boldsymbol{\mu}}, \mathbf{v}) = \frac{2}{\sum_{i=1}^{n}
				v_i} \sum_{i=1}^{n} v_i \left( \hat{\mu}_i - y_i + y_i
			\log\left(\nicefrac{y_i}{\hat{\mu}_i}\right)\right) \1{y_i > 0} + v_i
			\hat{\mu}_i \1{y_i = 0}.
		\end{equation*}
		In a claim frequency setting, $y_i$ denotes the observed frequency, and $v_i
			y_i$ denotes the claim count for policy $i$.
	\end{example}
	\noindent
\end{appendices}

%% file: my_bibliography.bib
@book{wuthrich_statistical_2023,
  address   = {Cham, Switzerland},
  series    = {Springer {Actuarial}},
  title     = {Statistical {Foundations} of {Actuarial} {Learning} and {Its} {Applications}},
  isbn      = {978-3-031-12408-2},
  language  = {eng},
  publisher = {Springer},
  author    = {Wüthrich, Mario V. and Merz, Michael},
  year      = {2023},
  doi       = {10.1007/978-3-031-12409-9}
}

@article{nelder_generalized_1972,
  title   = {Generalized {Linear} {Models}},
  volume  = {135},
  issn    = {0035-9238},
  doi     = {10.2307/2344614},
  number  = {3},
  urldate = {2023-09-20},
  journal = {Journal of the Royal Statistical Society. Series A (General)},
  author  = {Nelder, J. A. and Wedderburn, R. W. M.},
  year    = {1972},
  note    = {Publisher: [Royal Statistical Society, Wiley]},
  pages   = {370--384}
}

@misc{noll_case_2020,
  address    = {Rochester, NY},
  type       = {{SSRN} {Scholarly} {Paper}},
  title      = {Case {Study}: {French} {Motor} {Third}-{Party} {Liability} {Claims}},
  shorttitle = {Case {Study}},
  doi        = {10.2139/ssrn.3164764},
  language   = {en},
  urldate    = {2023-10-26},
  author     = {Noll, Alexander and Salzmann, Robert and Wuthrich, Mario V.},
  month      = mar,
  year       = {2020}
}

@manual{dutang_casdatasets_2018,
  title  = {{CASdatasets}: {Insurance} {Datasets}},
  author = {Dutang, Christophe and Charpentier, Arthur},
  year   = {2018},
  note   = {R package version 1.0–8},
  url    = {http://dutangc.free.fr/pub/RRepos/}
}

@inproceedings{sklearn_api,
  author    = {Lars Buitinck and Gilles Louppe and Mathieu Blondel and
               Fabian Pedregosa and Andreas Mueller and Olivier Grisel and
               Vlad Niculae and Peter Prettenhofer and Alexandre Gramfort
               and Jaques Grobler and Robert Layton and Jake VanderPlas and
               Arnaud Joly and Brian Holt and Ga{\"{e}}l Varoquaux},
  title     = {{API} {Design} for {Machine} {Learning} {Software}: {Experiences} from the {Scikit}-{Learn}
               {Project}},
  booktitle = {ECML PKDD Workshop: Languages for Data Mining and Machine Learning},
  year      = {2013},
  pages     = {108--122}
}

@article{LindholmW,
  title   = {The {Balance} {Property} in {Insurance} {Pricing}},
  author  = {M. Lindholm and Mario V. Wüthrich},
  journal = {Scandinavian Actuarial Journal},
  year    = {2025},
  volume  = {2025},
  number  = {},
  pages   = {},
  url     = {https://ssrn.com/abstract=4925165}
}

@article{holvoet_neural_2025,
  title      = {Neural {Networks} for {Insurance} {Pricing} with {Frequency} and {Severity} {Data}: {A} {Benchmark} {Study} from {Data} {Preprocessing} to {Technical} {Tariff}},
  volume     = {29},
  issn       = {1092-0277},
  shorttitle = {Neural {Networks} for {Insurance} {Pricing} with {Frequency} and {Severity} {Data}},
  doi        = {10.1080/10920277.2025.2451860},
  number     = {3},
  urldate    = {2025-09-19},
  journal    = {North American Actuarial Journal},
  author     = {Holvoet, Freek and Antonio, Katrien and Henckaerts, Roel},
  month      = jul,
  year       = {2025},
  pages      = {519--562}
}

@article{wuthrich_gini_2023,
  title    = {Model {Selection} with {Gini} {Indices} under {Auto}-calibration},
  volume   = {13},
  issn     = {2190-9741},
  doi      = {10.1007/s13385-022-00339-9},
  language = {en},
  number   = {1},
  urldate  = {2024-03-01},
  journal  = {European Actuarial Journal},
  author   = {Wüthrich, Mario V.},
  month    = jun,
  year     = {2023},
  pages    = {469--477}
}

@article{denuit_testing_2024,
  title    = {Testing for {Auto}-calibration with {Lorenz} and {Concentration} {Curves}},
  volume   = {117},
  issn     = {0167-6687},
  doi      = {10.1016/j.insmatheco.2024.04.003},
  urldate  = {2025-09-24},
  journal  = {Insurance: Mathematics and Economics},
  author   = {Denuit, Michel and Huyghe, Julie and Trufin, Julien and Verdebout, Thomas},
  month    = jul,
  year     = {2024},
  pages    = {130--139}
}

@inproceedings{schlimmer_beyond_1986,
  author    = {Schlimmer, Jeffrey C. and Granger, Richard H.},
  title     = {Beyond {Incremental} {Processing}: {Tracking} {Concept} {Drift}},
  year      = {1986},
  publisher = {AAAI Press},
  booktitle = {Proceedings of the Fifth AAAI National Conference on Artificial Intelligence},
  pages     = {502–507},
  numpages  = {6},
  location  = {Philadelphia, Pennsylvania},
  series    = {AAAI'86}
}

@article{widmer_learning_1996,
  title    = {Learning in the {Presence} of {Concept} {Drift} and {Hidden} {Contexts}},
  volume   = {23},
  issn     = {1573-0565},
  doi      = {10.1007/BF00116900},
  language = {en},
  number   = {1},
  urldate  = {2025-09-21},
  journal  = {Machine Learning},
  author   = {Widmer, Gerhard and Kubat, Miroslav},
  month    = apr,
  year     = {1996},
  pages    = {69--101}
}

@inproceedings{gama_learning_2004,
  address   = {Berlin, Heidelberg},
  title     = {Learning with {Drift} {Detection}},
  isbn      = {978-3-540-28645-5},
  doi       = {10.1007/978-3-540-28645-5_29},
  language  = {en},
  booktitle = {Advances in {Artificial} {Intelligence} – {SBIA} 2004},
  publisher = {Springer},
  author    = {Gama, João and Medas, Pedro and Castillo, Gladys and Rodrigues, Pedro},
  editor    = {Bazzan, Ana L. C. and Labidi, Sofiane},
  year      = {2004},
  pages     = {286--295}
}

@book{wasserman_all_2006,
  address   = {New York, NY},
  series    = {Springer {Texts} in {Statistics}},
  title     = {All of {Nonparametric} {Statistics}},
  copyright = {http://www.springer.com/tdm},
  isbn      = {978-0-387-25145-5},
  language  = {en},
  urldate   = {2025-09-28},
  publisher = {Springer},
  author    = {Wasserman, Larry},
  year      = {2006},
  doi       = {10.1007/0-387-30623-4}
}

@inproceedings{nishida_detecting_2007,
  address   = {Berlin, Heidelberg},
  title     = {Detecting {Concept} {Drift} {Using} {Statistical} {Testing}},
  isbn      = {978-3-540-75488-6},
  doi       = {10.1007/978-3-540-75488-6_27},
  language  = {en},
  booktitle = {Discovery {Science}},
  publisher = {Springer},
  author    = {Nishida, Kyosuke and Yamauchi, Koichiro},
  editor    = {Corruble, Vincent and Takeda, Masayuki and Suzuki, Einoshin},
  year      = {2007},
  pages     = {264--269}
}

@incollection{bifet_learning_2007,
  series    = {Proceedings},
  title     = {Learning from {Time}-{Changing} {Data} with {Adaptive} {Windowing}},
  isbn      = {978-0-89871-630-6},
  urldate   = {2025-09-23},
  booktitle = {Proceedings of the 2007 {SIAM} {International} {Conference} on {Data} {Mining} ({SDM})},
  publisher = {Society for Industrial and Applied Mathematics},
  author    = {Bifet, Albert and Gavaldà, Ricard},
  month     = apr,
  year      = {2007},
  doi       = {10.1137/1.9781611972771.42},
  pages     = {443--448}
}

@article{davidson_reliable_2009,
  title    = {Reliable {Inference} for the {Gini} {Index}},
  volume   = {150},
  issn     = {0304-4076},
  doi      = {10.1016/j.jeconom.2008.11.004},
  number   = {1},
  urldate  = {2025-09-24},
  journal  = {Journal of Econometrics},
  author   = {Davidson, Russell},
  month    = may,
  year     = {2009},
  pages    = {30--40}
}

@article{gama_survey_2014,
  title   = {A {Survey} on {Concept} {Drift} {Adaptation}},
  volume  = {46},
  issn    = {0360-0300},
  doi     = {10.1145/2523813},
  number  = {4},
  urldate = {2025-09-20},
  journal = {ACM Comput. Surv.},
  author  = {Gama, João and Žliobaitė, Indrė and Bifet, Albert and Pechenizkiy, Mykola and Bouchachia, Abdelhamid},
  month   = mar,
  year    = {2014},
  pages   = {44:1--44:37}
}

@article{rabanser_failing_2019,
  title   = {Failing {Loudly}: {An} {Empirical} {Study} of {Methods} for {Detecting} {Dataset} {Shift}},
  author  = {Rabanser, Stephan and G{\"u}nnemann, Stephan and Lipton, Zachary},
  journal = {Advances in Neural Information Processing Systems},
  volume  = {32},
  year    = {2019}
}

@article{denuit_model_2019,
  title    = {Model {Selection} {Based} on {Lorenz} and {Concentration} {Curves}, {Gini} {Indices} and {Convex} {Order}},
  volume   = {89},
  issn     = {0167-6687},
  doi      = {10.1016/j.insmatheco.2019.09.001},
  urldate  = {2025-09-25},
  journal  = {Insurance: Mathematics and Economics},
  author   = {Denuit, Michel and Sznajder, Dominik and Trufin, Julien},
  month    = nov,
  year     = {2019},
  pages    = {128--139}
}

@article{lu_learning_2019,
  title      = {Learning under {Concept} {Drift}: {A} {Review}},
  volume     = {31},
  issn       = {1558-2191},
  shorttitle = {Learning under {Concept} {Drift}},
  doi        = {10.1109/TKDE.2018.2876857},
  number     = {12},
  urldate    = {2025-09-21},
  journal    = {IEEE Transactions on Knowledge and Data Engineering},
  author     = {Lu, Jie and Liu, Anjin and Dong, Fan and Gu, Feng and Gama, João and Zhang, Guangquan},
  month      = dec,
  year       = {2019},
  pages      = {2346--2363}
}

@misc{ackerman_automatically_2021,
  title     = {Automatically {Detecting} {Data} {Drift} in {Machine} {Learning} {Classifiers}},
  doi       = {10.48550/arXiv.2111.05672},
  urldate   = {2025-09-20},
  publisher = {arXiv},
  author    = {Ackerman, Samuel and Raz, Orna and Zalmanovici, Marcel and Zlotnick, Aviad},
  month     = nov,
  year      = {2021}
}

@article{mallick_matchmaker_2022,
  title      = {Matchmaker: {Data} {Drift} {Mitigation} in {Machine} {Learning} for {Large}-scale {Systems}},
  volume     = {4},
  shorttitle = {Matchmaker},
  url        = {https://proceedings.mlsys.org/paper_files/paper/2022/hash/069a002768bcb31509d4901961f23b3c-Abstract.html},
  language   = {en},
  urldate    = {2025-09-21},
  journal    = {Proceedings of Machine Learning and Systems},
  author     = {Mallick, Ankur and Hsieh, Kevin and Arzani, Behnaz and Joshi, Gauri},
  month      = apr,
  year       = {2022},
  pages      = {77--94}
}

@article{hu_concept_2025,
  title    = {Concept {Drift} {Detection} {Based} on {Deep} {Neural} {Networks} and {Autoencoders}},
  volume   = {15},
  issn     = {2076-3417},
  doi      = {10.3390/app15063056},
  language = {en},
  number   = {6},
  urldate  = {2025-09-21},
  journal  = {Applied Sciences},
  author   = {Hu, Lisha and Lu, Yaru and Feng, Yuehua},
  month    = mar,
  year     = {2025},
  note     = {Publisher: MDPI AG},
  pages    = {3056}
}

@article{hinder_one_a_2024,
  title      = {One or {Two} {Things} we {Know} about {Concept} {Drift}—a {Survey} on {Monitoring} in {Evolving} {Environments}. {Part} {A}: {Detecting} {Concept} {Drift}},
  volume     = {7},
  issn       = {2624-8212},
  shorttitle = {One or {Two} {Things} we {Know} about {Concept} {Drift}—a {Survey} on {Monitoring} in {Evolving} {Environments}. {Part} {A}},
  doi        = {10.3389/frai.2024.1330257},
  language   = {English},
  urldate    = {2025-09-21},
  journal    = {Frontiers in Artificial Intelligence},
  author     = {Hinder, Fabian and Vaquet, Valerie and Hammer, Barbara},
  month      = jun,
  year       = {2024},
  note       = {Publisher: Frontiers}
}

@article{delong_isotonic_2025,
  title   = {Isotonic {Regression} for {Variance} {Estimation} and {Its} {Role} in {Mean} {Estimation} and {Model} {Validation}},
  volume  = {29},
  issn    = {1092-0277},
  doi     = {10.1080/10920277.2024.2421221},
  number  = {3},
  urldate = {2025-10-15},
  journal = {North American Actuarial Journal},
  author  = {Delong, Lukasz and Wüthrich, Mario V.},
  month   = jul,
  year    = {2025},
  pages   = {563--591}
}

@misc{brauer_wuethrich_gini_2025,
  title     = {Gini {Score} under {Ties} and {Case} {Weights}},
  doi       = {10.48550/arXiv.2511.15446},
  urldate   = {2025-11-25},
  publisher = {arXiv},
  author    = {Brauer, Alexej and Wüthrich, Mario V.},
  month     = nov,
  year      = {2025}
}

@article{wuthrich_isotonic_2024,
  title    = {Isotonic {Recalibration} under a {Low} {Signal}-to-noise {Ratio}},
  volume   = {2024},
  issn     = {0346-1238},
  doi      = {10.1080/03461238.2023.2246743},
  number   = {3},
  journal  = {Scandinavian Actuarial Journal},
  author   = {Wüthrich, Mario V. and Ziegel, Johanna},
  month    = mar,
  year     = {2024},
  pages    = {279--299}
}

@article{jorgensen_properties_1986,
  title   = {Some {Properties} of {Exponential} {Dispersion} {Models}},
  volume  = {13},
  issn    = {0303-6898},
  url     = {https://www.jstor.org/stable/4616024},
  number  = {3},
  urldate = {2025-10-24},
  journal = {Scandinavian Journal of Statistics},
  author  = {Jørgensen, Bent},
  year    = {1986},
  note    = {Publisher: [Board of the Foundation of the Scandinavian Journal of Statistics, Wiley]},
  pages   = {187--197}
}

@article{jorgensen_exponential_1987,
  title   = {Exponential {Dispersion} {Models}},
  volume  = {49},
  issn    = {0035-9246},
  url     = {https://www.jstor.org/stable/2345415},
  number  = {2},
  urldate = {2025-10-24},
  journal = {Journal of the Royal Statistical Society. Series B (Methodological)},
  author  = {Jørgensen, Bent},
  year    = {1987},
  note    = {Publisher: [Royal Statistical Society, Oxford University Press]},
  pages   = {127--162}
}

@article{murphy_new_1973,
  title    = {A {New} {Vector} {Partition} of the {Probability} {Score}},
  volume   = {12},
  issn     = {1520-0450},
  doi      = {10.1175/1520-0450(1973)012<0595:ANVPOT>2.0.CO;2},
  language = {EN},
  number   = {4},
  urldate  = {2025-10-28},
  journal  = {Journal of Applied Meteorology and Climatology},
  author   = {Murphy, Allan H.},
  month    = jun,
  year     = {1973},
  note     = {Publisher: American Meteorological Society
              Section: Journal of Applied Meteorology and Climatology},
  pages    = {595--600}
}

@article{delong_comparing_1988,
  title   = {Comparing the {Areas} under {Two} or {More} {Correlated} {Receiver} {Operating} {Characteristic} {Curves}: {A} {Nonparametric} {Approach}},
  volume  = {44},
  issn    = {0006-341X},
  doi     = {10.2307/2531595},
  number  = {3},
  urldate = {2025-11-10},
  journal = {Biometrics},
  author  = {DeLong, Elizabeth R. and DeLong, David M. and Clarke-Pearson, Daniel L.},
  year    = {1988},
  note    = {Publisher: [Wiley, International Biometric Society]},
  pages   = {837--845}
}

@article{gneiting_strictly_2007,
  title    = {Strictly {Proper} {Scoring} {Rules}, {Prediction}, and {Estimation}},
  volume   = {102},
  issn     = {0162-1459},
  doi      = {10.1198/016214506000001437},
  number   = {477},
  urldate  = {2025-11-25},
  journal  = {Journal of the American Statistical Association},
  author   = {Gneiting, Tilmann and Raftery, Adrian E},
  month    = mar,
  year     = {2007},
  pages    = {359--378}
}

@article{gneiting_making_2011,
  title    = {Making and {Evaluating} {Point} {Forecasts}},
  volume   = {106},
  issn     = {0162-1459},
  doi      = {10.1198/jasa.2011.r10138},
  number   = {494},
  urldate  = {2025-11-25},
  journal  = {Journal of the American Statistical Association},
  author   = {Gneiting, Tilmann},
  month    = jun,
  year     = {2011},
  pages    = {746--762}
}

@article{frees_insurance_2014,
  title   = {Insurance {Ratemaking} and a {Gini} {Index}},
  volume  = {81},
  issn    = {0022-4367},
  url     = {https://www.jstor.org/stable/24546807},
  number  = {2},
  urldate = {2025-12-05},
  journal = {The Journal of Risk and Insurance},
  author  = {Frees, Edward W. and Meyers, Glenn and Cummings, A. David},
  year    = {2014},
  pages   = {335--366}
}

@article{frees_summarizing_2011,
  title    = {Summarizing {Insurance} {Scores} {Using} a {Gini} {Index}},
  volume   = {106},
  issn     = {0162-1459},
  doi      = {10.1198/jasa.2011.tm10506},
  number   = {495},
  journal  = {Journal of the American Statistical Association},
  author   = {Frees, Edward W. and Meyers, Glenn and Cummings, A. David},
  month    = sep,
  year     = {2011},
  pages    = {1085--1098}
}
